\pdfoutput=1

\documentclass[11pt]{article}

\usepackage[preprint]{acl}

\usepackage{amsmath,amsfonts,bm}

\def\eqref#1{equation~\ref{#1}}

\def\1{\bm{1}}

\DeclareMathAlphabet{\mathsfit}{\encodingdefault}{\sfdefault}{m}{sl}
\SetMathAlphabet{\mathsfit}{bold}{\encodingdefault}{\sfdefault}{bx}{n}

\usepackage{times}
\usepackage{latexsym}

\usepackage[T1]{fontenc}

\usepackage[utf8]{inputenc}

\usepackage{microtype}

\usepackage{inconsolata}

\usepackage{graphicx}

\usepackage{xcolor}
\usepackage{color,colortbl}
\definecolor{darkgreen}{rgb}{0,0.5,0}
\definecolor{azureblue}{rgb}{0,0.5,1}
\definecolor{darkgreen}{rgb}{1,0,0}
\definecolor{color1}{HTML}{006EB8}
\definecolor{color2}{HTML}{009B55}
\definecolor{color3}{HTML}{00A99A}
\definecolor{color4}{HTML}{3C8031}
\definecolor{color5}{HTML}{006795}
\definecolor{color6}{HTML}{00AEB3}
\definecolor{mygray}{gray}{0.93}
\definecolor{mygreen}{HTML}{3FBC9D}
\definecolor{arsenic}{rgb}{0.23, 0.27, 0.29}

\usepackage{hyperref}
\hypersetup{colorlinks=true,urlcolor=color1,linkcolor=color1,citecolor=color1}
\usepackage{url}

\usepackage{tcolorbox}
\usepackage{graphicx}
\usepackage{wrapfig}
\usepackage{booktabs}
\usepackage{algorithm}
\usepackage{algpseudocode}
\usepackage{algorithmicx}
\usepackage{amsmath}
\usepackage{algpseudocode}
\usepackage{multirow}
\usepackage{multicol}
\usepackage{pifont}
\usepackage{xcolor,colortbl}
\usepackage{mathtools}
\usepackage{enumitem}
\usepackage{setspace}
\usepackage{fontawesome}
\usepackage{cleveref}
\usepackage{ulem}
\usepackage[T1]{fontenc}
\definecolor{pli-color}{HTML}{002FA7}
\definecolor{ured}{HTML}{EF426F}
\newcommand{\newtxt}[1]{\textcolor{black}{#1}}

\newcommand{\methodlong}[0]{\textbf{G}lobal and \textbf{L}ocal \textbf{I}nstruction \textbf{D}riven \textbf{E}xpert \textbf{R}outer}
\newcommand{\method}[0]{$\mathtt{GLIDER}$}

\title{Glider: Global and Local Instruction-Driven Expert Router} 

\author{Pingzhi Li$^{*}$\quad Prateek Yadav$^{*}$ \quad  Jaehong Yoon \quad  Jie Peng \quad Yi-Lin Sung \\\textbf{Mohit Bansal} \quad \textbf{Tianlong Chen}\\
The University of North Carolina at Chapel Hill \\
\texttt{\{pingzhi, praty\}@cs.unc.edu}}

\begin{document}
\maketitle

\def\thefootnote{*}\footnotetext{Equal contribution}

\begin{abstract}
The development of performant pre-trained models has driven the advancement of routing-based expert models tailored to specific tasks. However, these methods often favor generalization over performance on held-in tasks.
This limitation adversely impacts practical applicability, as real-world deployments require robust performance across both known and novel tasks. 
We observe that current token-level routing mechanisms neglect the global semantic context of the input task.
To address this, we propose a novel method, \methodlong{} (\method{}) that proposes a multi-scale routing mechanism, encompassing a semantic global router and a learned local router. 
The global router leverages recent LLMs' semantic reasoning capabilities to generate task-specific instructions from the input query, guiding expert selection across all layers. This global guidance is complemented by a local router that facilitates token-level routing decisions within each module, enabling finer control and enhanced performance on unseen and challenging tasks. 
Our experiments using T5-based expert models for T0 and FLAN tasks demonstrate that \method{} achieves substantially improved held-in performance while maintaining strong generalization on held-out tasks. Additionally, we perform ablations experiments to dive deeper into the components of \method{} and plot routing distributions to show that \method{} can effectively retrieve the correct expert for held-in tasks while also demonstrating compositional capabilities for held-out tasks. 
Our experiments highlight the importance of our multi-scale routing that leverages LLM-driven semantic reasoning for MoErging methods. 
Our code is available at \href{https://github.com/UNITES-Lab/glider}{https://github.com/UNITES-Lab/glider}. 

\end{abstract}

\begin{figure*}[t]
    \vspace{-20pt}
    \centering
    \includegraphics[width=1\textwidth]{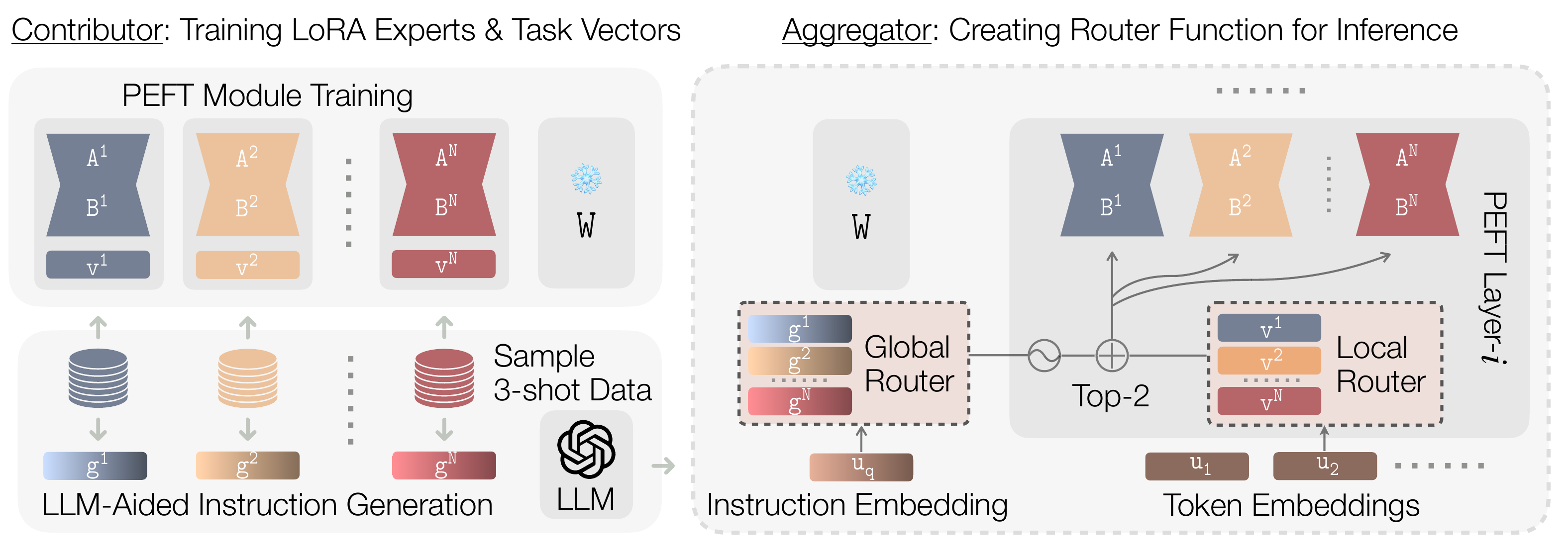}
    \vspace{-20pt}
    \caption{Overview of our method. \uline{Contributor}~(left): Each contributor utilizes local data to train several components: the PEFT module (comprising $\mathtt{A_i}$ and $\mathtt{B_i}$), task vectors ($\mathtt{v_i}$), and global routing vectors ($\mathtt{g_i}$). For the latter, an LLM is employed to generate semantically-informed instructions based on $3$ randomly selected examples, which are then embedded into $\mathtt{g_i}$.  \uline{Aggregator}~(right): The aggregator utilizes local and global task vectors to construct local routers $ [ \bar{\mathtt{v}}^\mathtt{1}; \hdots; \bar{\mathtt{v}}^\mathtt{N}]$ and a global router $[\mathtt{g}^\mathtt{1}; \hdots; \mathtt{g}^\mathtt{N}]$, respectively. For each query, the global router uses an LLM-generated instruction embedding to produce the global routing score. This score is then scaled and combined with the local routing score, enabling fine-grained control over expert selection.
    } 
    \vspace{-10pt}
    \label{fig:overview}
\end{figure*}

\section{Introduction}
\label{sec:introduction}

The emergence of highly capable large language models~(LLMs) has marked an increased attention in downstream task specialization. This specialization often leverages parameter-efficient fine-tuning (PEFT) techniques, such as LoRA~\citep{hu2021lora}, which introduce minimal trainable parameters (``adapters") to adapt pre-trained LLMs for specific tasks. The compact size of these specialized PEFT modules enables easy sharing, which has led to the distribution of an evergrowing number of adapters on various platforms. 

This proliferation of expert models, \textit{i.e.}~specialized adapters, has led to the development of methods for re-using such experts to improve performance or generalization~\citep{muqeeth2024learning,ostapenko2024towards,huang2024lorahub}. Central to these approaches are routing mechanisms that adaptively select relevant experts for a particular task or query. These routing methods have been referred to as ``Model MoErging”~\citep{yadav2024moergingsurvey} since they frequently share methodologies and ideas with mixture-of-experts (MoE) models~\citep{shazeer2017outrageously,fedus2022switch,du2022glam} and model merging~\citep{yadav2023ties,yadav2023compeft,ilharco2022editing}. However, MoE methods train experts jointly from scratch~\citep{gupta2022sparsely} while MoErging utilizes a decentralized, community-sourced pool of pre-trained experts. Furthermore, it departs from traditional model merging techniques by dynamically and adaptively combining these experts, optimizing performance at the query or task level.
MoErging methods offer three key advantages: ($1$) They support decentralized model development by reusing and routing among independently trained experts, reducing reliance on centralized resources. ($2$) They facilitate modular capability expansion and ``transparency" in updates as they either add or modify specialized expert models.
(3) They allow for compositional generalization by recombining fine-grained skills from various experts, extending the system's abilities to new unseen tasks beyond the capabilities of the individual expert models.


\newtxt{Most MoErging~\citep{chronopoulou2023adaptersoup,muqeeth2024learning,zhao2024loraretriever} methods prioritize either known or unseen tasks, limiting real-world applicability where both are critical. Real-world queries often span domains and defy clean categorization into predefined task boundaries. For instance, translating and analyzing text requires collaboration between multiple experts rather than selecting a single specialized model. Current approaches struggle in such scenarios. Phatgoose demonstrates this tradeoff, excelling on unseen tasks but underperforming on known ones.}

We hypothesize that this gap arises from the model's token-level routing mechanism. We show that for the held-in tasks, the independent routing decisions at each layer, based solely on individual token embeddings, lack sufficient global context to retrieve the correct expert for all tokens at every module. 
This leads to suboptimal routing, which may propagate noise through the network, further hindering accurate expert utilization in deeper layers. 
This highlights a critical limitation of token-level approaches to handling held-in tasks, which hence falls short of the goal of building a routing system that seamlessly handles arbitrary queries.
We believe that adding a global routing mechanism based on semantic task information can aid the token-level router for the correct retrieval of held-in tasks. Hence, we ask the question.

\begin{tcolorbox}[before skip=0.2cm, after skip=0.2cm, boxsep=0.0cm, middle=0.2cm, top=0.2cm, bottom=0.2cm]
\textit{\textbf{(Q)}} \textit{Can we leverage LLMs to generate semantics-aware task instructions to guide routing mechanism to facilitate both specialization and generalization?}
\end{tcolorbox}

This paper addresses the challenges by investigating the potential of leveraging the inherent reasoning and generalization capabilities of LLMs to guide the routing process in an MoE-like model composed of specialized LoRA modules. We introduce, \methodlong{} (\method{}) that hinges on a multi-scale routing mechanism that contains both local and global routers \newtxt{to select top-$2$ expert models} as shown in Figure~\ref{fig:overview}. The global router leverages LLM-generated, semantics-aware instructions (see Appendix~\ref{app:generated_instructions}) for each input query to score expert models. This high-level guidance is then complemented by a learned local router, which makes token-level routing decisions at each module, enabling fine-grained control and improving performance on the challenging held-out tasks. Through this framework, we highlight the crucial role of LLM reasoning in unlocking the compositional generalization capabilities of MoE models.

To test the effectiveness of our \method{} method, we follow Phatgoose~\citep{muqeeth2024learning} and use T5 models~\citep{raffel2020exploring} to create expert models for T0 held-in~\citep{sanh2022multitask} and FLAN tasks~\citep{longpre2023flan} and test performance on T0 held-in \& held-out~\citep{sanh2022multitask} and big-bench lite~\citep{srivastava2023beyond} \& hard tasks~\citep{suzgun2022challenging}. 
Our key contributions and findings are:

\begin{itemize}[nosep]
    \item We introduce \method{}, which employs LLM-guided multi-scale global and local attention. Our experiments show that \method{} outperforms previous methods, significantly improving performance on held-in tasks (\textit{e.g.} $6.6\%$ over Phatgoose on T0 held-in) while also enhancing zero-shot held-out compositional generalization (\textit{e.g.} $0.9\%$ on T0 held-out).
   \item We find that without LLM assistance, MoE models underperform individual specialized models on held-in tasks by $8.2\%$. Incorporating semantic-aware instructions enables \method{} to achieve comparable performance, demonstrating the LLM's capacity to effectively infer task identity and guide module selection without explicit task labels.
    \item \method{} also maintains strong performance on held-out tasks, showcasing its adaptability and generalization capabilities. Our work highlights the critical role of LLMs in enhancing MoE models' compositional generalization, advancing the development of more robust and versatile AI systems capable of handling both familiar and novel tasks.

\end{itemize}

\section{Related Works}

The abundance of specialized expert models has spurred the development of techniques to leverage ``experts" models for enhanced performance and generalization. \citet{yadav2024moergingsurvey} called such techniques as ``MoErging" \footnote{See e.g. \url{https://huggingface.co/spaces/open-llm-leaderboard/open_llm_leaderboard}} methods which rely on adaptive routing mechanisms to select relevant experts for specific tasks or queries. 
These methods can be broadly classified into four categories based on the design of their routing mechanisms.

\paragraph{Embedding-Based~Routing:} This category encompasses methods that derive routing decisions from learned embeddings of expert training data. These methods typically compare a query embedding against the learned expert embeddings to determine the optimal routing path.  Examples include AdapterSoup~\citep{chronopoulou2023adaptersoup}, Retrieval of Experts~\citep{jang2023exploring}, LoraRetriever~\citep{zhao2024loraretriever}, Mo'LoRA~\citep{maxine2023llama}, the embedding-based approach of Airoboros~\citep{durbin2024airoboros}, and Dynamic Adapter Merging~\citep{cheng2024dam}. 

\paragraph{Classifier-Based~Routing:} This category consists of methods that train a router to function as a classifier. This router is trained to predict the optimal routing path based on features extracted from expert datasets or unseen data.  Representative methods in this category include Zooter~\citep{lu2023routing}, Branch-Train-Mix~\citep{sukhbaatar2024branch}, Routing with Benchmark Datasets~\citep{shnitzer2023large}, Routoo~\citep{mohammadshahi2024leeroo}, and RouteLLM~\citep{ong2024routellmlearningroutellms}.  The key distinction between embedding-based and classifier-based routing lies in the router's architecture and training methodology. While embedding-based routing often employs a nearest neighbor approach, classifier-based routing typically relies on logistic regression or analogous classification techniques.

\paragraph{Task-Specific~Routing:} This category focuses on methods tailored to enhance performance on specific target tasks. These methods learn a task-specific routing distribution over the target dataset to optimize performance for the given task. Methods include LoraHub~\citep{huang2023lorahub}, LoRA-Flow~\citep{wang2024lora}, AdapterFusion~\citep{pfeiffer2020adapterfusion}, $\pi$-Tuning~\citep{wu2023pi}, Co-LLM~\citep{shen2024learning}, Weight-Ensembling MoE~\citep{tang2024wemoe}, MoLE~\citep{wu2024mole}, MeteoRA~\citep{Xu2024MeteoRAME}, PEMT~\citep{lin2024pemt}, MixDA~\citep{Diao2023MixtureofDomainAdaptersDA}, and Twin-Merging~\citep{lu2024twin}.

\paragraph{Routerless~Methods:} This final category encompasses methods that do not rely on an explicitly trained router. Instead, these methods often employ alternative mechanisms, such as heuristics or rule-based systems, for routing decisions. Examples include Arrow~\citep{ostapenko2024towards}, PHATGOOSE~\citep{muqeeth2024learning}, the ``ask an LLM" routing of Airoboros~\citep{durbin2024airoboros} and LlamaIndex~\citep{liu2024llamaindex}. \newtxt{Phatgoose and Arrow use only local routers, in contrast, \method{} uses both local and global guidance for routing.}

\begin{figure*}[htbp]
    \vspace{-20pt}
    \centering
    \includegraphics[width=0.98\textwidth]{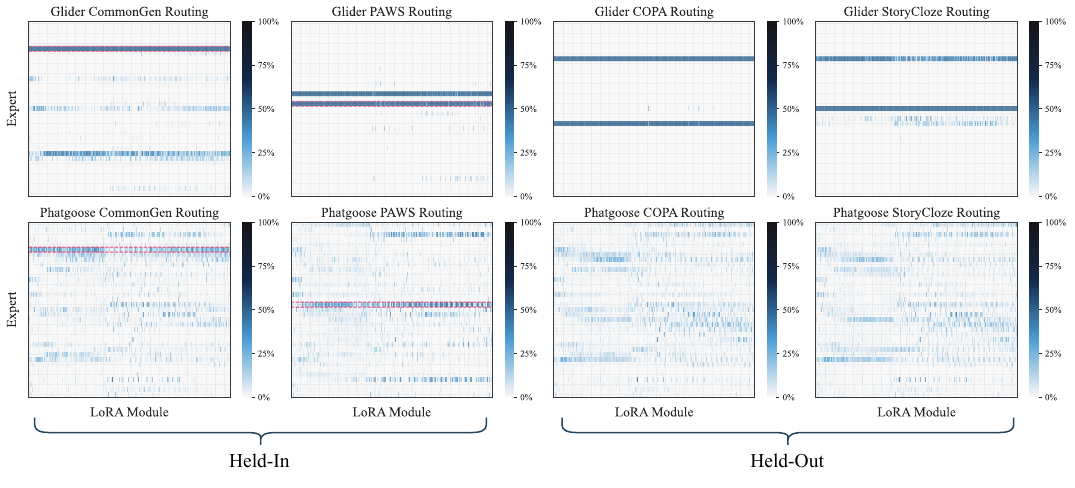}
    \vspace{-10pt}
    \caption{We present routing heatmaps for \method{} and Phatgoose on two held-in and two held-out tasks. For held-in tasks, oracle experts are marked with red dashed lines. \method{} selects oracle experts more frequently than Phatgoose for held-in tasks, leading to improvements of $3.3\%$ on CommonGen and $6.5\%$ on PAWS. For held-out tasks, \method{} also tends to select the most relevant experts across most LoRA modules, resulting in improvements of $2.2\%$ on COPA and $5.8\%$ on StoryCloze.}
    \vspace{-15pt}
    \label{fig:routing-heatmap}
\end{figure*}

\section{Problem Statement}
\label{sec:problem_statement}

In our work, we aim to build a routing mechanism capable of performing well on diverse queries from various tasks, including both seen and unseen tasks. For each query/token and module, this routing mechanism dynamically selects a model from a large pool of specialized expert models to achieve high performance. 
To facilitate modular development, we adopt a \textit{contributor-aggregator} framework~\citep{yadav2024moergingsurvey} where individual contributors create specialized expert models from a generalist model for their respective tasks and distribute these models to others for public usage. The aggregator builds a routing mechanism over the expert models that shared by the contributor to direct queries to the most relevant experts. 
Following recent works~\citep{muqeeth2024learning,ostapenko2024towards}, we use parameter-efficient finetuning (PEFT)~\citep{liu2022few,Sung2022LST,poth2023adapters} methods like LoRA~\citep{lora} to train the expert models. Since PEFT typically has lower computational and communication costs than full-model finetuning~\citep{lora,liu2022few}, the use of PEFT makes it easier to participate and contribute. PEFT methods introduce modules throughout the model – for example, LoRA~\citep{lora} introduces a low-rank update at every linear layer in the model. We refer to each of these updates as a \textit{module}. 
Subsequently, the trained expert models and additional information are shared with the aggregators. The aggregator's job is to collect these expert models and the additional information and design the post-hoc routing mechanism. This mechanism will effectively direct incoming queries to the most appropriate expert model for each token and at each module to ensure optimal performance on both seen and unseen tasks. This approach allows for the seamless integration of new capabilities by adding expert models to the existing pool. Next, we formally define our contributor-aggregator framework.

Let us assume that there are $N$ contributors, $\{c_1, c_2, \hdots, c_\mathtt{N}\}$, and each contributor $c_i$ has access to a task-specific datasets $\mathcal{D}_i$. Each contributor, $c_i$, follows the predefined training protocol $\mathcal{T}$ provided by the aggregator.
The training protocol ($\mathcal{T}$) takes in a base model ($\boldsymbol{\theta}_\textrm{base}$) and a dataset ($\mathcal{D}_i$). It returns the expert model parameters ($\phi_{i}$) along with any additional information ($\Psi_{i}$) that needs to be shared with the aggregators, for example, the gate vectors described in \Cref{ssec:expert_training_protocol}. Specifically, $\{\phi_{i},~\Psi_{i}\} \leftarrow \mathcal{T}(\boldsymbol{\theta}_\textrm{base}, \mathcal{D}_i)$.  
All contributors share this information with the aggregator, which creates a pool of models containing $\{(\phi_i, \Psi_i)\}_{i=1}^\mathtt{N}$. The aggregators ($\mathcal{A}$) then uses these expert models and the auxiliary information to create a routing mechanism $\mathcal{R}(.)$ that takes the user query $q$ as the input and return routing path describing how the information is routed through the given set of expert models. Formally, $\mathcal{R}(.) \leftarrow \mathcal{A}(\{(\phi_i, \Psi_i)\}_{i=1}^\mathtt{N})$. The function $\mathcal{R}(.)$ describe the full path of input query by making various choices about 1) expert input granularity, choosing to route per-token, per-query, or per-task, 2) expert depth granularity, opting for either per-module or model-level routing, and 3) selecting between sparse or dense routing. Finally, the aggregator uses the routing mechanism to answer incoming queries.

\section{Methodology}
\label{sec:method}

To recap, our goal is to build a MoErging method that dynamically routes queries to a diverse pool of specialized expert models, addressing the challenge of effectively handling queries from various tasks and ensuring both held-in and held-out performance. Our proposed method, \methodlong{} (\method{}), leverages a combination of local and global routing vectors to achieve this goal. Specifically, contributors train task-specific routing vectors, while an LLM generates global semantic task instructions, which are then converted to global instruction routing vectors.  During inference, these local and global routing vectors are combined to perform top-k discrete routing, directing queries to the most suitable expert model. This process is visualized in Figure~\ref{fig:overview} and described in detail below.

\subsection{Expert Training Protocol} \label{ssec:expert_training_protocol}

Our expert training protocol $\mathcal{T}$ takes as input the base model parameters, $\theta_\textrm{base}$, and a dataset $\mathtt{d}$ and performs three steps to obtain the required output. First, we train the LoRA experts ($\phi$) and then the local routing vectors ($\mathtt{l}$) while keeping the LoRA experts fixed. Finally, we train the global routing vector ($\mathtt{g}$) by using an LLM and an embedding model. Formally, in our case, $\phi,~\Psi = \{\mathtt{l}, \mathtt{g}\} \leftarrow \mathcal{T}(\theta_\textrm{base}, \mathtt{d})$ which are then shared with the aggregators to create the routing mechanism. We described these steps in detail below.

\paragraph{PEFT Training of Expert Model.} 
\method{} is compatible with expert models trained using parameter-efficient finetuning methods (\textit{e.g.} LoRA~\citep{lora}, Adapters~\citep{houlsby2019parameter}) that introduce small trainable modules throughout the model. We focus on PEFT experts because they typically have lower computational and communication costs than full-model finetuning~\citep{yadav2023compeft}, making it easier to train and share expert models. Following Phatgoose~\citep{muqeeth2024learning}, this work specifically focuses on LoRA~\citep{lora} due to its widespread use. 
LoRA introduces a \underline{\textit{module}} comprising the trainable matrices $\mathtt{B}\in \mathbb{R}^{\mathtt{d} \times \mathtt{r}}$ and $\mathtt{A} \in \mathbb{R}^{\mathtt{r} \times \mathtt{n}}$ in parallel to each linear layer with parameters $\mathtt{W} \in \mathbb{R}^{\mathtt{d} \times \mathtt{n}}$. Given the $\mathtt{i}^{\text{th}}$ input token activation $\mathtt{u_i}$, LoRA modifies the output of the linear layer from $\mathtt{Wu_i}$ to $\mathtt{Wu_i + \frac{\alpha}{r} \cdot BAu_i}$ where $\mathtt{\alpha}$ is a constant and usually is set to $\mathtt{1}$. During training, the matrices $\mathtt{A}$ and $\mathtt{B}$ are trainable, while the original linear layer $\mathtt{W}$ is kept frozen. We denote the final trained expert parameters with $\phi = \{(\mathtt{A_1}, \mathtt{B_1}), \hdots, (\mathtt{A_m}, \mathtt{B_m})\}$, where $\mathtt{m}$ is the number of modules in the model.

\paragraph{Training Local Routing Vectors.} 

Following Phatgoose~\citep{muqeeth2024learning}, after training the PEFT modules on their dataset, a local router is introduced before each PEFT module. This router, employing a shared vector across all queries and tokens, dynamically determines the utilization of the PEFT module based on the input token activations. The router is trained for a small number of steps using the same dataset and objective as the PEFT module while keeping the expert PEFT parameters fixed. This process effectively learns to associate the token activation patterns with the learned expert model. For LoRA, the local router, represented by a trainable vector  $\mathtt{v} \in \mathbb{R}^{\mathtt{d}}$, controls the contribution of the PEFT module to the final output. This results in a modified linear layer of the form $\mathtt{Wu_i + \frac{\alpha}{r} \cdot BAu_i \cdot\newtxt{\texttt{sigmoid}}(v^\mathsf{T}u_i)}$, where $\mathtt{\alpha}$, $\mathtt{W}$, $\mathtt{B}$, and $\mathtt{A}$ are frozen, and the local router $\mathtt{v}$ is learned. We denote the final local routing vectors as $\mathtt{l} = \{\mathtt{v_1}, \hdots, \mathtt{v_m}\}$ where $\mathtt{m}$ is the number of modules in the model.

\paragraph{Creating LLM-Aided Global Routing Vector.} 
The local routing vectors capture the intricate relationships between token activations and expert models, enabling efficient query routing in cases where no dedicated expert is available. Conversely, for queries corresponding to held-in tasks, direct retrieval of the relevant expert model is preferred to process the full query. For this purpose, we create a global routing vector that utilizes an LLM to generate a semantically-informed instruction, termed as \underline{task description}, which effectively captures the essence of the kind of queries the expert can handle. We prompt an LLM with three randomly selected in-context examples to generate this task description. We used the \texttt{gpt-4-turbo} model along with the prompt provided in Appendix~\ref{sec:task_description}. The resulting task description is then embedded using an off-the-shelf embedding model, specifically the \texttt{nomic-embed-text-v1.5} model, to produce a global routing vector for the task. We denote the global routing vector as $\mathtt{g} \in \mathbb{R}^{\mathtt{d_g}}$.

\subsection{\method{}: Inference Expert Aggregation}

Following training, all contributors share their expert models along with the auxiliary information comprising of the local and global routing vectors, $\{\mathtt{\phi^t},~\mathtt{l^t},~\mathtt{g^t}\}_{\mathtt{t=1}}^{\mathtt{N}}$, \newtxt{where t indexes the input tokens} with the aggregators. The \method{} method subsequently leverages this information to perform inference on arbitrary queries.

\paragraph{Local Router.} Before each input module $\mathtt{m}$, a separate local router \newtxt{weight} $\mathtt{L}_{\mathtt{m}} \in \mathbb{R}^{\mathtt{N \times d}}$ is inserted to make local per-token, per-module routing decisions. 
For a given module $\mathtt{m}$ and expert model $\mathtt{c}$, \newtxt{we have $\bar{\mathtt{v}}_\mathtt{m}^\mathtt{c} = \frac{\mathtt{v_m^c} - \mu(\mathtt{v_m^c})}{\sigma(\mathtt{v_m^c})}$, where $\mu(\cdot)$ and $\sigma(\cdot)$ denote the mean and standard deviation respectively.}
Next, we obtain the local router for module $\mathtt{m}$ by stacking these standardised local routing vectors as $\mathtt{L}_{\mathtt{m}} = [ \bar{\mathtt{v}}_\mathtt{m}^\mathtt{1}; \hdots; \bar{\mathtt{v}}_\mathtt{m}^\mathtt{N}] \in \mathbb{R}^{\mathtt{N \times d}}$. Next, for each token $\mathtt{i}$ with activation $\mathtt{u_i}$ coming into module $\mathtt{m}$, we standardise it to obtain $\bar{\mathtt{u}}_\mathtt{i} = \frac{\mathtt{u}_\mathtt{i} - \mu(\mathtt{u}_\mathtt{i})}{\sigma{(\mathtt{u}_\mathtt{i}})}$. We then compute the local affinity scores, $\mathtt{s}^\mathtt{loc}_{\mathtt{m}} \in \mathbb{R}^{\mathtt{N}}$ between the local router $\mathtt{L}_{\mathtt{m}}$ and $\mathtt{u_i}$ as $\mathtt{s}^\mathtt{loc}_{\mathtt{m}} = \texttt{cos-sim}(\mathtt{L}_{\mathtt{m}}, \mathtt{u_i})$.

\paragraph{Global Router.} 
The global router aims to capture task semantics to retrieve relevant experts for any given input query. We create the global router \newtxt{weight} $\mathtt{G} \in \mathbb{R}^{\mathtt{N \times d_g}}$ by stacking the global routing vectors from all the expert models as $\mathtt{G} = [\mathtt{g}^\mathtt{1}; \hdots; \mathtt{g}^\mathtt{N}]$.  This router is not a part of the base model and is added before the model to independently process the full query. Given an input query $\mathtt{u}$ along with three few-shot input-output pairs of similar queries, we prompt an LLM (\texttt{gpt-4-turbo}) using the template provided in Appendix~\ref{sec:task_description} to obtain a task description for the query. We then embed this task description using the same embedding model (\texttt{nomic-embed-text-v1.5}) to obtain the vector $\mathtt{q_u} \in \mathbb{R}^{\mathtt{d_g}}$. We then compute the global affinity score, $\mathtt{s}^{\mathtt{glob}} \in \mathbb{R}^{\mathtt{N}}$, by computing the cosine similarity as $\mathtt{s}^{\mathtt{glob}} = \texttt{cos-sim}(\mathtt{G}, \mathtt{q_u})$.

\paragraph{Combining Global and Local Router.} At each module $\mathtt{m}$, we have the global and local affinity scores $\mathtt{s}^\mathtt{glob}$ and $\mathtt{s}^\mathtt{loc}_{\mathtt{m}}$ respectively. Following Phatgoose~\citep{muqeeth2024learning}, we scale the local scores with a factor of $1/\sqrt{\mathtt{N}}$. However, the global router's main goal is to retrieve the correct expert for the held-in tasks. Therefore, we first check if the expert with the highest global affinity score ($\texttt{max}(\mathtt{s}^\mathtt{glob})$) is above a threshold ($\mathtt{p}$). If such experts exist, then we set a high $\alpha$ to enforce retrieval and vice versa. Hence, we propose to scale the global scores with $\alpha$, where $\alpha = \gamma \cdot\mathbb{I}_{\{\texttt{max}(\mathtt{s}^\mathtt{glob}) - \mathtt{p} > 0\}} + \beta$, where $\mathtt{p}$ is the cosine similarity threshold, and $\gamma$ and $\beta$ are scaling hyperparameters. Using our ablation experiments in Section~\ref{sec:ablation}, we set $\mathtt{p}=\mathtt{0.8}$, $\gamma = 100$ and $\beta=3$. 
We then obtain the final affinity score $\mathtt{s} \in \mathbb{R}^{\mathtt{N}} = \alpha \cdot\mathtt{s}^\mathtt{glob}$ + $\mathtt{s}^\mathtt{loc}_{\mathtt{m}} / \sqrt{\mathtt{N}}$. Then \method{} selects the top-$\mathtt{k}$ experts after performing \texttt{softmax} over the final affinity score $\mathtt{s}$ as $\mathcal{E}_{\mathtt{top}}$ = \texttt{top}-$\mathtt{k}(\texttt{softmax}(\mathtt{s}))$. 
Finally, the output of the module for token activation $u_i$ is computed as $\mathtt{Wu_i} + \sum_{\mathtt{k} \in \mathcal{E}_\mathtt{top}} \mathtt{s_{k}\cdot B_kA_ku_i}$.

\section{Experiments}

\subsection{Setting}

\paragraph{Dataset.}

Our experiments utilize the multitask prompted training setup introduced by \citet{sanh2021multitask}, which has become a standard benchmark for evaluating \newtxt{held-in performance as well as} generalization to unseen tasks \citep{chung2022scaling,longpre2023flan,jang2023exploring,zhou2022not}. 
\newtxt{Phatgoose~\citep{muqeeth2024learning} show how local routing can be used for generalization to unseen domain, hence,
following them,} we employ LM-adapted T5.1.1 XL~\citep{lester2021power} as our base model which is a 3B parameter variant of T5~\citep{raffel2020exploring} further trained on the C4 dataset using a standard language modeling objective. For held-out evaluations, we follow Phatgoose~\citep{muqeeth2024learning} and use three held-out benchmark collections. We use the T0 held-out (T0HO) datasets used in~\citet{sanh2021multitask} and the two subsets of BIG-bench~\citep{srivastava2023beyond}. Specifically, we use BIG-bench Hard (BBH)~\citep{suzgun2022challenging}, consisting of 23 challenging datasets, and BIG-bench Lite (BBL)~\citep{srivastava2023beyond}, a lightweight 24-dataset proxy for the full benchmark. Similar to \citet{muqeeth2024learning}, we exclude certain BIG-bench datasets due to tokenization incompatibility with the T5 tokenizer.

\paragraph{Expert Creation.}
To create the pool of expert module for routing, we follow \citet{muqeeth2024learning} and use two distinct dataset collections: \ding{182} T0 Held-In~\citep{sanh2021multitask} consisting of the 36 held-in prompted datasets for tasks from the T0 training procedure. \ding{183} The ``FLAN Collection"~\citep{longpre2023flan} which significantly expands the T0 tasks by incorporating prompted datasets from SuperGLUE~\citep{wang2019superglue}, Super Natural Instructions~\citep{wang2022super}, dialogue datasets, and Chain-of-Thought datasets~\citep{wei2022chain}. Following \citet{muqeeth2024learning}, we create $\mathtt{166}$ specialized models from the FLAN Collection. For each dataset in these collections, we train Low-Rank Adapters (LoRAs)~\citep{hu2021lora} modules resulting in pools of $\mathtt{36}$ and $\mathtt{166}$ expert models for T0 Held-In and FLAN, respectively. Similar to Phatgoose, we use a rank of $r = 16$ and train for $\mathtt{1000}$ steps using the AdamW optimizer~\citep{Loshchilov2017DecoupledWD} with a learning rate of $5 \times 10^{-3}$ and a warmup ratio of $0.06$. After training the LoRA module, we freeze it and train the local routing vectors for an additional 100 steps with the same hyperparameters. Finally, following prior work~\citep{shazeer2016outrageously,du2022glam,lepikhin2020gshard}, \method{} performs top-$k$ routing with $k = 2$.

\begin{table}[t]
    \vspace{-20pt}
  \centering
  \caption{Performance evaluated on the T0 set and FLAN set. We present the performance on both held-in tasks (\textit{i.e.} T0-HI) and held-out tasks (\textit{i.e.} T0-HO, BBH, and BBL). We compare the following methods: ($1$) performance upper bound, \textit{i.e.} Oracle Expert; ($2$) zero-shot baselines, \textit{i.e.} Multi-Task Fine-Tuning, Expert Merging, Arrow, and Phatgoose;  ($3$) few-shot baselines, \textit{i.e.} LoRA Hub and \method{}. We mark the best performance besides the upper bound (\textit{i.e.}, Oracle Expert) in \textbf{bold}.}
   \resizebox{0.98\linewidth}{!}{
    \begin{tabular}{l|cccc|cc}
    \toprule
    \midrule
    \multirow{2}[2]{*}{\textbf{Method}} & \multicolumn{4}{c|}{\textbf{T0}} & \multicolumn{2}{c}{\textbf{FLAN}} \\
    & \textbf{T0-HI} & \textbf{T0-HO} & \textbf{BBH} & \textbf{BBL} & \textbf{BBH} & \textbf{BBL}\\
    \midrule
    Oracle Expert & \textcolor{gray}{$69.60$} & \textcolor{gray}{$51.60$} & \textcolor{gray}{$34.90$} & \textcolor{gray}{$36.60$} & \textcolor{gray}{$38.90$} & \textcolor{gray}{$45.40$} \\
    \midrule
    Multi-Task Fine-Tuning & $55.90$ & $51.60$ & $34.90$ & $36.60$ & $\mathbf{38.90}$ & $\mathbf{45.40}$ \\
    Expert Merging & $30.73$ & $45.40$ & $\mathbf{35.30}$ & $36.00$ & $34.60$ & $34.00$ \\
    Arrow & $39.84$ & $55.10$ & $33.60$ & $34.50$ & $30.60$ & $29.60$ \\
    Phatgoose & $61.42$ & $56.90$ & $34.90$ & $37.30$  & $35.60$ & $35.20$ \\
    \midrule
    LoRA Hub & $31.90$ & $46.85$ &
    $31.35$ & 
    $31.18$ & $34.50$ & $30.54$   \\
    \method{} & $\mathbf{68.04}$ & $\mathbf{57.78}$ & $35.29$ & $\mathbf{37.46}$ & $
    35.07$ & $35.52$  \\
    \midrule
    \bottomrule
    \end{tabular}}
    \vspace{-10pt}
  \label{tab:main-results}
\end{table}

\subsection{Baselines}

\textbf{Expert Merging.}
Model Merging~\citep{yadav2023ties,choshen2022fusing} involves averaging the parameters of multiple models or modules to create a single aggregate model. We merge by multiplying the LoRA matrices and then taking an unweighted average of all the experts within the pool. It is important to note that this merging strategy requires homogeneous expert module architectures; in contrast, \method{} can accommodate heterogeneous expert modules.

\textbf{Arrow.} 
Following \citet{ostapenko2024towards}, we employ a routing mechanism where gating vectors are derived from LoRA expert modules.  Specifically, the first right singular vector of the outer product of each module's LoRA update ($BA$) serves as its gating vector. Input routing is determined by a probability distribution based on the absolute dot product between the input representation and each gating vector. We utilize top-$k$ routing with $k=2$.

\textbf{Phatgoose.} Phatgoose~\citep{muqeeth2024learning} first learn the LoRA modules for each, followed by learning a sigmoid gating vector similar to our local router. During inference, they make routing decisions for each token independently for all modules. Specifically, they first standardize the input token activations and gating vectors from all experts and then perform similarity-based top-2 routing.

\textbf{LoRA Hub.}
LoraHub~\citep{huang2023lorahub} method performs gradient-free optimization using few-shot task samples to learn mixing coefficients for different expert models while keeping them fixed. Once the coefficients are learned, they merge the experts with the learned weight and route through the merged expert.

\begin{figure}
    \vspace{-20pt}
    \centering
    \includegraphics[width=1.0\linewidth]{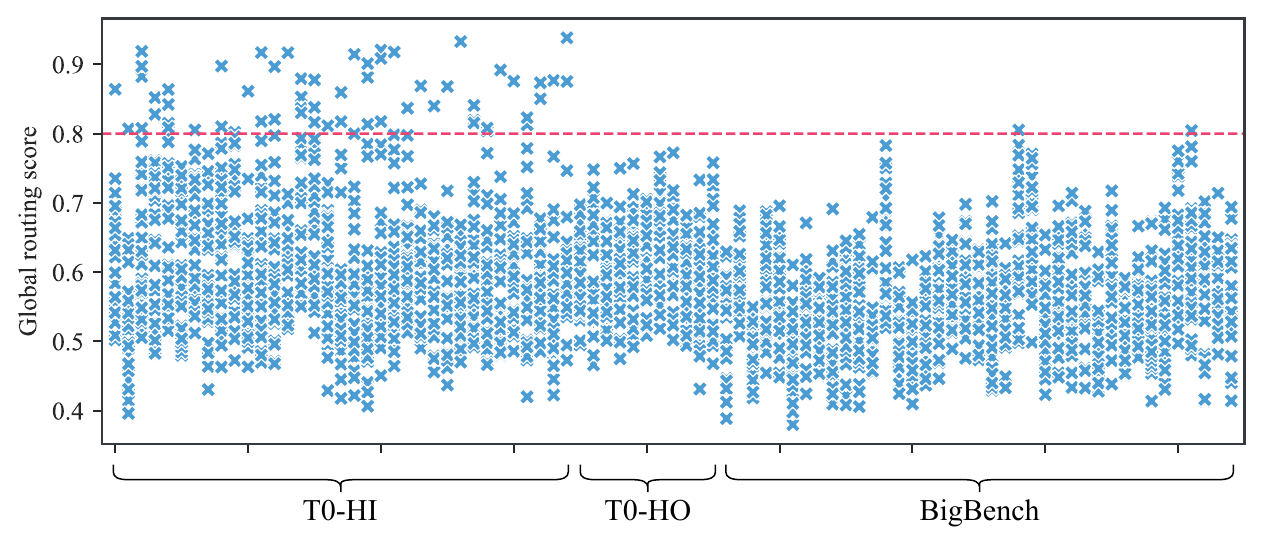}
    \vspace{-20pt}
    \caption{ Global routing scores for tasks in the T0 set. The red horizontal line indicates our design threshold of $0.8$. Each column represents an evaluated task from {T0-HI, T0-HO, BigBench} using T0 held-in experts. All global routing scores for each task are plotted, corresponding to the $35$ experts in total.}
    \label{fig:global-routing}
\end{figure}

\textbf{Multi-task Fine-Tuning.} 
Multitask training is a proven method for enhancing zero-shot generalization~\citep{sanh2021multitask,wei2021finetuned} but is infeasible given our problem setting and data access limitations. We include it as a baseline using publicly available models. Specifically, we utilize the T0-3B model~\citep{sanh2021multitask} for the T0 Held-In datasets, given its training on a matching dataset collection. For FLAN, a directly comparable publicly available model is unavailable; therefore, we report FLAN-T5 XL results trained on a different, undisclosed dataset mixture, while acknowledging the limitations of this indirect comparison.

\textbf{Oracle.} 
Following \cite{jang2023exploring} and \cite{muqeeth2024learning}, we employ an Oracle routing scheme as a performance upper bound. This scheme selects the expert exhibiting optimal performance on a given evaluation dataset, thus representing a non-zero-shot approach.

\subsection{Main Results}

Table~\ref{tab:main-results} presents the comparison results among our \method{} and six baselines on both held-in and held-out settings. \newtxt{We report the average performance across all tasks for each setting, please see Appendix~\ref{app:dataset} for each tasks metric.} To further illustrate the performance, we also include the results of Oracle Expert, which has extra access to the task identities of expert modules and evaluated datasets and can be regarded as an \textit{upper bound}.

\paragraph{T0 Setting.} In the T0 task set, the following observations can be drawn: \ding{182} For the held-in tasks, \textit{i.e.} T0-HI, \method{} significantly outperforms other baselines and almost matches the performance of Oracle Expert upper bound. \ding{183} For T0-HO and BBL tasks, \method{} achieves the best performance among all the methods, including Oracle Expert upper bound. \ding{184} \method{} has negligible lower performance, \textit{i.e.} $0.01\%$, compared to the Expert Merging baseline in BBH but outperforms it by around $12\%$ on T0-HO and $1.5\%$ on BBL. Besides Expert Merging, \method{} outperforms all other methods on BBH, including the Oracle Expert upper bound.

\begin{table}[t]
      \centering
      \caption{Ablation on the instruction coefficient $\alpha$. We mark the best performance in \textbf{bold} and the performance corresponding to the selected $\alpha$ by \method{} in \colorbox{cyan!30}{blue}. }
       \resizebox{0.7\linewidth}{!}{
        \begin{tabular}{l|cccc}
        \toprule
        \midrule
        \multirow{2}[2]{*}{\textbf{$\alpha$}} & \multicolumn{4}{c}{\textbf{T0}} \\
        & \textbf{T0-HI} & \textbf{T0-HO} & \textbf{BBH} & \textbf{BBL} \\
        \midrule
        $0$ & $61.42$ & $56.90$ & $34.90$ & $37.30$ \\
        $1$ & $62.20$ & $57.04$ & $35.05$ & $\mathbf{37.79}$   \\
        $3$ & $63.40$ & \colorbox{cyan!30}{$57.78$} & \colorbox{cyan!30}{$\mathbf{35.29}$} & \colorbox{cyan!30}{$37.46$}  \\
        $10$ & $65.52$ & $\mathbf{57.98}$ & $34.80$ & $37.04$    \\
        $100$ & \colorbox{cyan!30}{$\mathbf{68.04}$} & $53.22$ & $31.73$ & $34.97$    \\
        $1000$ & $66.88$ & $52.91$ & $30.71$ & $34.31$    \\
        $3000$ & $66.69$ & $52.37$ & $30.03$ & $33.24$    \\
        \midrule
        \bottomrule
        \end{tabular}}
        \vspace{-15pt}
      \label{tab:instruction-coefficient}
\end{table}

\subsection{Ablation Study and Further Investigation}
\label{sec:ablation}

\paragraph{Ablation on the global routing scale $\alpha$.} To illustrate how the specialization and generalization abilities change as we scale the coefficient $\alpha$ of the global routing score, we conduct the ablation study of $\alpha$ ranging $\{1, 3, 10, 100, 1000, 3000\}$. As shown in Table~\ref{tab:instruction-coefficient}, we present experimental results of the T0 task set on both held-in and held-out tasks. For held-in tasks, \textit{i.e.}~T0-HI, \method{} can select the optimal $\alpha$ to scale the global routing score. For held-out tasks, \textit{i.e.}~\{T0-HO, BBH, BBL\}, \method{} produce either the optimal $\alpha$~(for BBH) or the sub-optimal $\alpha$ with slightly lower performance to the optimal ones~(for T0-HO and BBL). \newtxt{Lastly, note that Phatgoose correspond to the setting where there is no global semantics used, i.e., $\alpha = 0$.}

\paragraph{Ablation on the routing strategy.} There exists a trade-off between performance and efficiency when using different $\mathtt{top}\texttt{-}\mathtt{k}$ routing strategies~\citep{ramachandran2018diversity}. To investigate the impact of routing strategy in \method{}, we evaluate $\mathtt{top}\texttt{-}\mathtt{k}$ routing of $\mathtt{k}$ in $\{1, 2, 3\}$. Moreover, we further evaluate the $\mathtt{top}\texttt{-}\mathtt{p}$ routing~\citep{huang2024hardertasksneedexperts,zeng2024adamoetokenadaptiveroutingnull} of $\mathtt{p}$ in $\{25\%, 50\%, 75\%\}$, where each token selects experts with higher routing probabilities until the cumulative probability exceeds threshold $\mathtt{p}$. As shown in Table~\ref{tab:top-k-routing-ablation}, we can draw the following conclusions: ($1$)~For $\mathtt{top}\texttt{-}\mathtt{k}$ routing, $\mathtt{k}=2$ shows comparable or better performance than $\mathtt{k}=3$, particularly for T0-HO and BBH, while offering improved efficiency. ($2$)~For $\mathtt{top}\texttt{-}\mathtt{p}$ routing, higher $\mathtt{p}$ values consistently yield better performance at the cost of efficiency. Therefore, we use $\mathtt{top}\texttt{-}\mathtt{2}$ routing in \method{} by default.

\begin{table}[t]
      \centering
      \caption{Ablation on the routing strategy. \method{} employs \colorbox{gray!20}{$\mathtt{top}\texttt{-}\mathtt{2}$} routing. We mark the best performance among  $\mathtt{top}\texttt{-}\mathtt{k}$ and $\mathtt{top}\texttt{-}\mathtt{p}$ routing in \textbf{bold}, respectively.}
       \resizebox{0.7\linewidth}{!}{
        \begin{tabular}{c|cccc}
        \toprule
        \midrule
        \multirow{2}[2]{*}{\textbf{Method}} & \multicolumn{4}{c}{\textbf{T0}} \\
        & \textbf{T0-HI} & \textbf{T0-HO} & \textbf{BBH} & \textbf{BBL} \\
        \midrule
        \texttt{Top-$1$} & $67.96$ & $56.07$ & $33.91$ & $35.82$ \\
        \rowcolor[gray]{0.9}
        \texttt{Top-$2$} & $68.04$ & $\mathbf{57.78}$ & $\mathbf{35.39}$ & $37.46$  \\
        \texttt{Top-$3$} & $\mathbf{68.06}$ & $57.52$ & $35.08$ & $\mathbf{38.55}$  \\
        \midrule
        \texttt{Top-$25\%$} & $67.98$ & $56.53$ & $34.10$ & $36.32$  \\
        \texttt{Top-$50\%$} & $67.95$ & $57.25$ & $35.07$ & $37.49$  \\
        \texttt{Top-$75\%$} & $\mathbf{68.02}$ & $\mathbf{57.86}$ & $\mathbf{35.38}$ & $\mathbf{38.65}$  \\
        \midrule
        \bottomrule
        \end{tabular}}
        \vspace{-20pt}
      \label{tab:top-k-routing-ablation}
    
\end{table}

\paragraph{Investigation on the threshold design of global scores.} As in Section~\ref{sec:method}, we compute the scale $\alpha$ for global scores using the formula $\alpha = \gamma * \mathbb{I}_{\{\texttt{max}(\mathtt{s}^\mathtt{glob}) - 0.8 > 0\}} + \beta$, where we establish a threshold of $0.8$ to differentiate evaluated tasks. Figure~\ref{fig:global-routing} presents the global routing scores for each task in the T0 set to motivate the rationale behind this design. For all \uline{held-in} tasks~(\textit{i.e.}, T0-HI), at least one expert~(typically the oracle expert trained on the evaluated task) achieves global routing scores exceeding $0.8$. Consequently, \method{} applies a higher $\alpha=100$, enabling effective identification of tasks corresponding to a specifically trained expert and enhancing retrieval of this oracle expert. For nearly all \uline{held-out} tasks~(\textit{i.e.}, T0-HO and BigBench), no global routing score surpasses $0.8$, prompting \method{} to utilize a lower $\alpha=3$. Two exceptions among the held-out tasks are \texttt{bbq\_lite\_json} and \texttt{strange\_stories} in BigBench, where one score marginally exceeds $0.8$ in each case. For these two, \method{} employs the higher $\alpha=100$, resulting in performance improvements of $1.3\%$ and $2.9\%$ respectively over $\alpha=3$, thus showing the effectiveness of our design.

\section{Conclusion}
\label{sec:conclusion}
This paper introduces \method{}, a novel multi-scale routing mechanism that incorporates both global semantic and local token-level routers. By leveraging the semantic reasoning capabilities of LLMs for global expert selection and refining these choices with a learned local router, \method{} addresses the limitations of existing methods that often perform poorly on held-in tasks. Our empirical evaluation on T0 and FLAN benchmarks, using T5-based experts, demonstrates that \method{} achieves substantial improvements in held-in task performance while maintaining competitive generalization on held-out tasks. These findings suggest that incorporating global semantic task context into routing mechanisms is crucial for building robust and practically useful routing-based systems.

\section{\newtxt{Limitation}}
\newtxt{The main limitation of \method{} lies in its heavy dependence on large language models (specifically GPT-4) for generating semantic task descriptions. This reliance introduces potential accessibility barriers due to API costs. Furthermore, investigating the application of \method{} to other modalities beyond language tasks, such as vision or multi-modal expert models, could unlock new capabilities for specialized model routing.} 

\bibliography{custom, datasets}

\appendix
\newpage
\appendix
\section*{Appendix}

\section{Extended Related Work}
\label{app:related_work}
\paragraph{Model Merging.} Model merging~\citep{yadav2023ties,choshen2022fusing,wortsman2022model,rame2022recycling,matena2022merging,ilharco2022editing,tam2023merging,jin2022dataless,yang2023adamerging,zhao2024model} consolidates multiple independently trained models with identical architectures into a unified model that preserves multi-model capabilities. While simple parameter averaging suffices for models within a linearly connected low-loss parameter space~\citep{mcmahan2017communication,stich2018local,frankle2020linear, wortsman2021learning,li2023merge}, more sophisticated techniques are necessary for complex scenarios. For instance, task vectors facilitate merging expert models trained on diverse domains~\citep{ilharco2022editing}. Additionally, methods like weighted merging using Fisher Importance Matrices~\citep{matena2022merging, tam2023merging} and TIES-Merging, which addresses sign disagreements and redundancy~\citep{yadav2023ties} offers improved performance. As a non-adaptive expert aggregation method, merging serves as a fundamental baseline for numerous Model Editing with Regularization (MoErging) techniques.

\paragraph{Multitask Learning (MTL)} research offers valuable insights for decentralized development. Notably, investigations into task-relatedness~\citep{standley2020tasks, bingel2017identifying, achille2019task2vec, vu2020exploring, Zamir2018TaskonomyDT, Mou2016HowTA} provide guidance for designing routing mechanisms, while MTL architectures addressing the balance between shared and task-specific knowledge~\citep{Misra2016CrossStitchNF, Ruder2017LatentMA, Meyerson2017BeyondSH, Zaremoodi2018AdaptiveKS, Sun2019AdaShareLW} offer strategies for combining expert contributions in a decentralized manner.

\paragraph{MoE for Multitask Learning.} Recent research has extensively investigated mixture-of-experts (MoE) models for multitask learning, achieving promising results in unseen task generalization. These approaches generally fall into two categories: ($1$) Example Routing: Studies like \citet{muqeeth2023soft, zadouri2023pushing, wang2022adamix} train routers to dynamically select experts for each input, while \citet{caccia2023multi} demonstrate the efficacy of routing at a finer granularity by splitting expert parameters into blocks.  ($2$) Task Routing:  \citet{ponti2023combining} employs a trainable skill matrix to assign tasks to specific parameter-efficient modules, while \citet{gupta2022sparsely} leverages task-specific routers selected based on domain knowledge. \citet{ye2022eliciting} proposes a layer-wise expert selection mechanism informed by task representations derived from input embeddings. Such approaches leverage task-specific representation to allow the router to effectively select the most suitable experts for unseen tasks. While these studies differ from our setting by assuming simultaneous data access, they offer valuable insights applicable to our exploration of  creating routing mechanisms over expert models.

\section{LLM for Task Instruction Generation.}
\label{sec:task_description}

\subsection{Prompt Template}

We use the following prompt with $3$ randomly selected samples for each task to generate its description. The prompt is then fed into the \texttt{gpt-4-turbo} OpenAI API to get the generated task descriptions.

\begin{tcolorbox}[before skip=0.2cm, after skip=0.2cm, boxsep=0.0cm, middle=0.2cm, top=0.2cm, bottom=0.2cm]
\textit{The following are three pairs of input-output examples from one task. Generate the task instruction in one sentence that is most possibly used to command a language model to produce them. In the instruction, remember to point out the skill or knowledge required for the task to guide the language model. 
\newline
\newline
- Input:
\newline
- Output: 
\newline
\newline
- Input: 
\newline
- Output:
\newline
\newline
- Input:
\newline
- Output:}
\end{tcolorbox}

\subsection{Examples of the Generated Instructions}
\label{app:generated_instructions}
We provide several examples of LLM-generated instructions in this section.

\textbf{WikiBio}~\citep{lebret2016neuraltextgenerationstructured} (T0 Held-In):
\begin{itemize}
    \item \textit{Create a short biography using the provided facts, demonstrating knowledge in historical and biographical writing.}
    \item \textit{Write a short biography based on the given factual bullet points, demonstrating proficiency in summarizing and transforming structured data into coherent narrative text.}
\end{itemize}

\textbf{CommonGen}~\citep{lin2020commongenconstrainedtextgeneration} (T0 Held-In):
\begin{itemize}
    \item \textit{Generate a coherent sentence using all the given abstract concepts, requiring the skill of concept integration to form a meaningful sentence.}
    \item \textit{Generate a coherent sentence by creatively combining a given set of abstract concepts.}
\end{itemize}

\textbf{COPA}~\citep{huang2024copageneralroboticmanipulation} (T0 Held-Out):
\begin{itemize}
    \item \textit{Identify the most logically consistent sentence from two given options based on the provided context, demonstrating reasoning and causal relationship skills.}
    \item \textit{Generate the most likely outcome for a given scenario by choosing between two provided options based on contextual clues and causal reasoning.}
\end{itemize}

\textbf{Date Understanding}~\citep{srivastava2023imitationgamequantifyingextrapolating} (BigBench-Hard):
\begin{itemize}
    \item \textit{Calculate the date based on the given information and present it in MM/DD/YYYY format, ensuring that you accurately account for day, month, and year changes.}
\end{itemize}

\textbf{Hindu Mythology Trivia}~\citep{srivastava2023imitationgamequantifyingextrapolating} (BigBench-Lite):
\begin{itemize}
    \item \textit{Generate the correct answer by making use of your knowledge in Hindu mythology and culture.}
\end{itemize}

\section{Demonstrating Compositional Generation}

In addition to significant improvements on held-in tasks, \method{} demonstrates strong performance on held-out tasks, showcasing its generalization capability. To further examine this ability to handle unseen tasks by composing experts, we provide specific task examples illustrating the association between selected experts and the evaluated task. As Figure~\ref{fig:routing-heatmap} shows, \method{} primarily selects two experts for the COPA (T0 held-out) task, corresponding to CosmosQA and QuaRel. The following three examples from these tasks demonstrate their close semantic relationship:

\begin{itemize}
    \item \textbf{COPA}: 
    \begin{itemize}
        \item \uline{Question:} \textit{Everyone in the class turned to stare at the student.
Select the most plausible cause: 
- The student's phone rang.
- The student took notes.} 
        \item \uline{Answer:} \textit{The student's phone rang.}
    \end{itemize}
    \item \textbf{CosmosQA:} 
     \begin{itemize}
        \item \uline{Question:} \textit{That idea still weirds me out . I made a blanket for the baby 's older sister before she was born but I completely spaced that this one was on the way , caught up in my own dramas and whatnot . Luckily , I had started a few rows in white just to learn a stitch ages ago , and continuing that stitch will make an acceptable woobie , I think .
According to the above context, choose the best option to answer the following question.
Question: What did I make for the baby .
Options:
A. I made a carseat .
B. None of the above choices .
C. I made a crb .
D. I finished a pair of booties .} 
        \item \uline{Answer:} \textit{D.}
    \end{itemize}
    \item \textbf{QuaRel:}
    \begin{itemize}
        \item \uline{Question:} \textit{Here's a short story: A piece of thread is much thinner than a tree so it is (A) less strong (B) more strong.
What is the most sensical answer between "Thread" and  "Tree"?}
        \item \uline{Answer:} \textit{Thread.}
    \end{itemize}
\end{itemize}

\color{black}
\section{Datasets and Metric}
\label{app:dataset}

\newtxt{The specific details of all the datasets we use in this work are provided in this section.}

\subsection{\newtxt{T0 Held-In Datasets}}

\begin{itemize}
    \item \textbf{CommonsenseQA}~\cite{talmor-etal-2019-commonsenseqa} under \textit{MIT License}, evaluated by accuracy.
    \item \textbf{DREAM}~\cite{sundream2018} under \textit{MIT License}, evaluated by accuracy.
    \item \textbf{QUAIL}~\cite{rogers2020getting} under \textit{CC BY-SA 4.0}, evaluated by accuracy.
    \item \textbf{QuaRTz}~\cite{quartz} under \textit{Apache 2.0 License}, evaluated by accuracy.
    \item \textbf{Social IQA}~\cite{sap-etal-2019-social} under \textit{MIT License}, evaluated by accuracy.
    \item \textbf{WiQA}~\cite{tandon-etal-2019-wiqa} under \textit{Apache 2.0 License}, evaluated by accuracy.
    \item \textbf{Cosmos QA}~\cite{huang-etal-2019-cosmos} under \textit{MIT License}, evaluated by accuracy.
    \item \textbf{QASC}~\cite{allenai:qasc} under \textit{Apache 2.0 License}, evaluated by accuracy.
    \item \textbf{Quarel}~\cite{quarel_v1} under \textit{Apache 2.0 License}, evaluated by accuracy.
    \item \textbf{SciQ}~\cite{SciQ} under \textit{MIT License}, evaluated by accuracy.
    \item \textbf{Wiki Hop}~\cite{welbl2018constructing} under \textit{CC BY-SA 3.0}, evaluated by accuracy.
    \item \textbf{Adversarial QA}~\cite{bartolo2020beat} under \textit{Apache 2.0 License}, evaluated by F1 score.
    \item \textbf{Quoref}~\cite{quoref} under \textit{Apache 2.0 License}, evaluated by F1 score.
    \item \textbf{DuoRC}~\cite{DuoRC} under \textit{MIT License}, evaluated by F1 score.
    \item \textbf{ROPES}~\cite{ropes} under \textit{Apache 2.0 License}, evaluated by F1 score.
    \item \textbf{Hotpot QA}~\cite{hotpotqa} under \textit{CC BY-SA 4.0}, evaluated by exact match and F1 score.
    \item \textbf{Wiki QA}~\cite{YangYihMeek:EMNLP2015:WikiQA} under \textit{MIT License}, evaluated by mean average precision (MAP) and mean reciprocal rank (MRR).
    \item \textbf{Common Gen}~\cite{lin-etal-2020-commongen} under \textit{MIT License}, evaluated by BLEU and ROUGE scores.
    \item \textbf{Wiki Bio}~\cite{wikibio} under \textit{CC BY-SA 3.0}, evaluated by BLEU score.
    \item \textbf{Amazon}~\cite{amazon} under \textit{Proprietary License}, evaluated by accuracy.
    \item \textbf{App Reviews}~\cite{appreviews} under \textit{Proprietary License}, evaluated by accuracy.
    \item \textbf{IMDB}~\cite{imdb} under \textit{Proprietary License}, evaluated by accuracy.
    \item \textbf{Rotten Tomatoes}~\cite{rottentomatoes} under \textit{Proprietary License}, evaluated by accuracy.
    \item \textbf{Yelp}~\cite{yelp} under \textit{Apache 2.0 License}, evaluated by accuracy.
    \item \textbf{CNN Daily Mail}~\cite{cnndailymail} under \textit{Apache 2.0 License}, evaluated by ROUGE score.
    \item \textbf{Gigaword}~\cite{graff2003english} under \textit{LDC License}, evaluated by ROUGE score.
    \item \textbf{MultiNews}~\cite{multinews} under \textit{MIT License}, evaluated by ROUGE score.
    \item \textbf{SamSum}~\cite{samsum} under \textit{CC BY-SA 4.0}, evaluated by ROUGE score.
    \item \textbf{XSum}~\cite{xsum} under \textit{Apache 2.0 License}, evaluated by ROUGE score.
    \item \textbf{AG News}~\cite{agnews} under \textit{CC BY-SA 3.0}, evaluated by accuracy.
    \item \textbf{DBPedia}~\cite{dbpedia} under \textit{CC BY-SA 3.0}, evaluated by accuracy.
    \item \textbf{TREC}~\cite{trec} under \textit{NIST License}, evaluated by accuracy.
    \item \textbf{MRPC}~\cite{dolan2005automatically} under \textit{Apache 2.0 License}, evaluated by accuracy and F1 score.
    \item \textbf{PAWS}~\cite{paws2019naacl} under \textit{Apache 2.0 License}, evaluated by accuracy and F1 score.
    \item \textbf{QQP}~\cite{qqp} under \textit{Quora Terms of Service}, evaluated by accuracy and F1 score.
\end{itemize}

\subsection{\newtxt{T0 Held-Out Datasets}}

\section*{Held-out Tasks}
\begin{itemize}
    \item \textbf{ANLI}~\cite{nie2019adversarial} under \textit{MIT License}, evaluated by accuracy.
    \item \textbf{CB}~\cite{de2019commitmentbank} under \textit{CC-BY-SA License}, evaluated by accuracy.
    \item \textbf{RTE}~\cite{dagan2005pascal, bar2006second, giampiccolo2007third, bentivogli2009fifth} under \textit{Apache 2.0 License}, evaluated by accuracy.
    \item \textbf{WSC}~\cite{levesque2012winograd} under \textit{Creative Commons License}, evaluated by accuracy.
    \item \textbf{Winogrande}~\cite{sakaguchi2019winogrande} under \textit{Apache License 2.0}, evaluated by accuracy.
    \item \textbf{WiC}~\cite{pilehvar2019wic} under \textit{CC BY-SA 4.0 License}, evaluated by accuracy.
    \item \textbf{COPA}~\cite{roemmele2011choice} under \textit{BSD-2-Clause License}, evaluated by accuracy.
    \item \textbf{HellaSwag}~\cite{zellers2019hellaswag} under \textit{MIT License}, evaluated by accuracy.
    \item \textbf{Story Cloze}~\cite{mostafazadeh2016corpus} under \textit{CC-BY 4.0 License}, evaluated by accuracy.
\end{itemize}

\subsection{\newtxt{BigBench-Hard Datasets}}

\begin{itemize}
    \item \textbf{abstract\_narrative\_understanding}~\cite{ghosh2021epic,Holyoak:2012,Nippold:2001,10.5555/2886521.2886662,wang-etal-2020-continuity,https://doi.org/10.48550/arxiv.1604.01696} under \textit{Apache 2.0 License}, evaluated by accuracy
    \item \textbf{abstraction\_and\_reasoning\_corpus}~\cite{chollet2019measure,Brown2020,AbstractionAndReasoningChallenge} under \textit{Apache 2.0 License}, evaluated by accuracy
    \item \textbf{anachronisms}~\cite{otterbacher-etal-2002-revisions,popescu-strapparava-2015-semeval,llorens-etal-2015-semeval,meng-etal-2017-temporal,Geva2021Aristotle} under \textit{Apache 2.0 License}, evaluated by accuracy
    \item \textbf{analogical\_similarity}~\cite{plate2003holographic,plate1994thesis,THAGARD1990259,GENTNER1993524} under \textit{Apache 2.0 License}, evaluated by accuracy
    \item \textbf{analytic\_entailment}~\cite{hume1739humannature,kant1781critique,wittgenstein1953,Quine1951trends,GriceStrawson1956,bolukbasi2016man,Kocurek2020against,Rudolph2020comparing,Kocurek2021counterlogicals} under \textit{Apache 2.0 License}, evaluated by accuracy
    \item \textbf{arithmetic}~\cite{Brown2020,Saxton2019analysing} under \textit{Apache 2.0 License}, evaluated by accuracy
    \item \textbf{ascii\_word\_recognition}~\cite{Child2019generating,Chen2020generative} under \textit{Apache 2.0 License}, evaluated by accuracy
    \item \textbf{authorship\_verification}~\cite{Bischoff2020,Koppel2004authorship} under \textit{Apache 2.0 License}, evaluated by accuracy
    \item \textbf{auto\_categorization} under \textit{Apache 2.0 License}, evaluated by accuracy
    \item \textbf{bbq\_lite}~\cite{Crawford2017trouble,khashabi-etal-2020-unifiedqa,li-etal-2020-unqovering} under \textit{Apache 2.0 License}, evaluated by accuracy
    \item \textbf{bias\_from\_probabilities}~\cite{10.1145/3442188.3445922,Abid2021persistent} under \textit{Apache 2.0 License}, evaluated by accuracy
    \item \textbf{boolean\_expressions}~\cite{habernal-etal-2018-argument,Yu2020ReClor,dua-etal-2019-drop,Liu2020LogiQA,Sinha2019CLUTRR,Wang2019SuperGLUEAS,Steinbach2002boolean,Saxton2019analysing,Payani2019learning,Trask2018units,Selsam2018SAT,Allamanis2016learning,Evans2018entailment,Shi2020reasoning} under \textit{Apache 2.0 License}, evaluated by accuracy
    \item \textbf{bridging\_anaphora\_resolution\_barqa}~\cite{hou-2020-bridging,hou-etal-2013-global,markert-etal-2012-collective,rajpurkar-etal-2016-squad} under \textit{Apache 2.0 License}, evaluated by accuracy
    \item \textbf{causal\_judgment}~\cite{Gordon2010COPA,Bosselut2019COMET,Halpern2016causality,Knobe2003intentional} under \textit{Apache 2.0 License}, evaluated by accuracy
    \item \textbf{cause\_and\_effect}~\cite{Gordon2010COPA} under \textit{Apache 2.0 License}, evaluated by accuracy
    \item \textbf{checkmate\_in\_one}~\cite{Alexander2020chess,Ammanabrolu2019quest,Dambekodi2020,Ammanabrolu2020stories} under \textit{Apache 2.0 License}, evaluated by accuracy
    \item \textbf{chess\_state\_tracking}~\cite{Weston2015toy,Cote2018TextWorld,Toshniwal2021blindfolded,Alexander2020chess,Chen2020transformers,noever2020chess,Swingle2021ChePT} under \textit{Apache 2.0 License}, evaluated by accuracy
    \item \textbf{chinese\_remainder\_theorem} under \textit{Apache 2.0 License}, evaluated by accuracy
    \item \textbf{cifar10\_classification} under \textit{Apache 2.0 License}, evaluated by accuracy
    \item \textbf{codenames}~\cite{Kim_Ruzmaykin_Truong_Summerville_2019} under \textit{Apache 2.0 License}, evaluated by accuracy
    \item \textbf{color}~\cite{Gibson2017color,Zucconi2016colour} under \textit{Apache 2.0 License}, evaluated by accuracy
    \item \textbf{com2sense}~\cite{singh-etal-2021-com2sense} under \textit{Apache 2.0 License}, evaluated by accuracy
    \item \textbf{common\_morpheme}~\cite{Devlin2018BERT,Wu2016Google,Won2021embedding,xu-etal-2018-incorporating,edmiston-stratos-2018-compositional,El-Kishky2019parsimonious} under \textit{Apache 2.0 License}, evaluated by accuracy
    \item \textbf{context\_definition\_alignment}~\cite{Senel2021wink,Reimers2019sentence} under \textit{Apache 2.0 License}, evaluated by accuracy
    \item \textbf{convinceme}~\cite{Lin2021Truthful,Levy2021memorization,Clark2018ARC,Maynez2020faithfulness,wang-etal-2020-asking,kenton2021alignment,Xu2020Recipes,Tamkin2021understanding} under \textit{Apache 2.0 License}, evaluated by accuracy
    \item \textbf{coqa\_conversational\_question\_answering}~\cite{reddy-etal-2019-coqa,radford2019language,Brown2020} under \textit{Apache 2.0 License}, evaluated by accuracy
    \item \textbf{crash\_blossom} under \textit{Apache 2.0 License}, evaluated by accuracy
    \item \textbf{crass\_ai}~\cite{Scholkopf2021causal,Teney2020learning,liang-etal-2020-learning,Pearl2000causality,Shahid2021counterfactual,Xia2021causal,Priol2020analysis} under \textit{Apache 2.0 License}, evaluated by accuracy
    \item \textbf{cryobiology\_spanish}~\cite{Scudellari2017,SoltaniFirouz2021,Toubiana2020,MBOGBA201895} under \textit{Apache 2.0 License}, evaluated by accuracy
    \item \textbf{cryptonite}~\cite{efrat-etal-2021-cryptonite,raganato-etal-2017-word,Sakaguchi2020WINOGRANDE,miller-gurevych-2015-automatic,miller-etal-2017-semeval,Joshi2018sarcasm,oprea-magdy-2020-isarcasm,Friedlander2018penny,Lewis2020question} under \textit{Apache 2.0 License}, evaluated by accuracy
    \item \textbf{cs\_algorithms} under \textit{Apache 2.0 License}, evaluated by accuracy
    \item \textbf{cycled\_letters}~\cite{Brown2020} under \textit{Apache 2.0 License}, evaluated by accuracy
    \item \textbf{dark\_humor\_detection}~\cite{weller-seppi-2019-humor,FAN2020105,Willinger2017demands,yang-etal-2015-humor,mihalcea-strapparava-2005-making} under \textit{Apache 2.0 License}, evaluated by accuracy
    \item \textbf{date\_understanding}~\cite{vashishth-etal-2018-dating,chambers-2012-labeling,Kotsakos2014burstiness,vashishtha-etal-2020-temporal,Vashishtha2019fine} under \textit{Apache 2.0 License}, evaluated by accuracy
    \item \textbf{disambiguation\_qa}~\cite{zhao-etal-2018-gender,rudinger-etal-2018-gender} under \textit{Apache 2.0 License}, evaluated by accuracy
    \item \textbf{discourse\_marker\_prediction}~\cite{malmi-etal-2018-automatic,nie-etal-2019-dissent,sileo-etal-2019-mining,sileo-etal-2020-discsense} under \textit{Apache 2.0 License}, evaluated by accuracy
    \item \textbf{disfl\_qa}~\cite{gupta-etal-2021-disfl,rajpurkar-etal-2018-know} under \textit{Apache 2.0 License}, evaluated by accuracy
    \item \textbf{diverse\_social\_bias}~\cite{sheng-etal-2019-woman,Nadeem2020StereoSet,Hendrycks2020aligning,sap-etal-2020-social,gehman-etal-2020-realtoxicityprompts,bolukbasi2016man,Caliskan2017semantics,may-etal-2019-measuring,liang-etal-2020-towards,Barocas2016big,cho-etal-2019-measuring,blodgett-etal-2020-language,Merity2016pointer,socher-etal-2013-recursive,poria-etal-2019-meld} under \textit{Apache 2.0 License}, evaluated by accuracy
    \item \textbf{dyck\_languages}~\cite{CHOMSKY1959118,Suzgun2019memory,Hao2018transductions,Hewitt2020RNNs,Hahn2020limitations,suzgun-etal-2019-lstm,Sennhauser2018evaluating,skachkova-etal-2018-closing,bhattamishra-etal-2020-ability,yu-etal-2019-learning,Ebrahimi2020Dyck,Ackerman2020survey,bhattamishra-etal-2020-practical} under \textit{Apache 2.0 License}, evaluated by accuracy
    \item \textbf{dynamic\_counting}~\cite{suzgun-etal-2019-lstm,skachkova-etal-2018-closing,bhattamishra-etal-2020-ability,Suzgun2019memory,yu-etal-2019-learning,Ebrahimi2020Dyck,Ackerman2020survey,bhattamishra-etal-2020-practical,Sennhauser2018evaluating,Merrill2020capacity,Karpathy2015} under \textit{Apache 2.0 License}, evaluated by accuracy
    \item \textbf{elementary\_math\_qa}~\cite{amini-etal-2019-mathqa,Ling2017induction,Hendrycks2021mathematical,patel-etal-2021-nlp,Zhang2020parsing,hendryckstest2021} under \textit{Apache 2.0 License}, evaluated by accuracy
    \item \textbf{emojis\_emotion\_prediction}~\cite{shoeb-de-melo-2020-emotag1200,PLUTCHIK19803} under \textit{Apache 2.0 License}, evaluated by accuracy
    \item \textbf{empirical\_judgments}~\cite{kant1781critique,kant1783prolegomena,Spirtes2000causation,Pearl1988probabalistic,Goldberger1972structural,Rothman2005causation,Roemmele2011COPA,Wang2019SuperGLUEAS,Evans2019sense} under \textit{Apache 2.0 License}, evaluated by accuracy
    \item \textbf{english\_proverbs}~\cite{GyasiObeng1996,Honeck1997proverb,HrisztovaGotthardt2015} under \textit{Apache 2.0 License}, evaluated by accuracy
    \item \textbf{english\_russian\_proverbs}~\cite{Bodrova2007Russian,Gvarjalaze1971,WikiquoteRussianProverbs} under \textit{Apache 2.0 License}, evaluated by accuracy
    \item \textbf{entailed\_polarity}~\cite{karttunen-2012-simple} under \textit{Apache 2.0 License}, evaluated by accuracy
    \item \textbf{entailed\_polarity\_hindi}~\cite{karttunen-2012-simple} under \textit{Apache 2.0 License}, evaluated by accuracy
    \item \textbf{epistemic\_reasoning}~\cite{sep-folkpsych-theory,Call2008chimpanzee,Bugnyar2016Ravens,Stalnaker1978assertion,Sperber2002pragmatics,nematzadeh-etal-2018-evaluating,le-etal-2019-revisiting,jiang-de-marneffe-2019-know,ross-pavlick-2019-well,bowman-etal-2015-large,de-marneffe-etal-2012-happen,jeretic-etal-2020-natural} under \textit{Apache 2.0 License}, evaluated by accuracy
    \item \textbf{evaluating\_information\_essentiality}~\cite{rajpurkar-etal-2018-know,hosseini-etal-2014-learning,levy-etal-2017-zero,Yin2015optimizing,de-marneffe-etal-2008-finding,zhu-etal-2019-learning} under \textit{Apache 2.0 License}, evaluated by accuracy
    \item \textbf{fact\_checker}~\cite{thorne-etal-2018-fever,lee-etal-2021-towards} under \textit{Apache 2.0 License}, evaluated by accuracy
    \item \textbf{factuality\_of\_summary}~\cite{eyal-etal-2019-question,wang-etal-2020-asking,durmus-etal-2020-feqa,vasilyev-etal-2020-fill,see-etal-2017-get,DBLP:conf/nips/HermannKGEKSB15,narayan-etal-2018-dont,pagnoni-etal-2021-understanding,kryscinski-etal-2020-evaluating} under \textit{Apache 2.0 License}, evaluated by accuracy
    \item \textbf{fantasy\_reasoning}~\cite{Wang2019SuperGLUEAS,wang-etal-2018-glue,McCann2018decathlon,Bhagavatula2019abductive,Lourie2021unicorn,Liu2020LogiQA,Saxton2019analysing,Clark2018ARC,Yu2019ActivityNet,Zhang2018record,Johnson2016CLEVR} under \textit{Apache 2.0 License}, evaluated by accuracy
    \item \textbf{few\_shot\_nlg}~\cite{Rastogi_Zang_Sunkara_Gupta_Khaitan_2020,kale-rastogi-2020-template} under \textit{Apache 2.0 License}, evaluated by accuracy
    \item \textbf{figure\_of\_speech\_detection}~\cite{Potamias2012transformer} under \textit{Apache 2.0 License}, evaluated by accuracy
    \item \textbf{forecasting\_subquestions} under \textit{Apache 2.0 License}, evaluated by accuracy
    \item \textbf{gem}~\cite{gehrmann-etal-2021-gem} under \textit{Apache 2.0 License}, evaluated by accuracy
    \item \textbf{gender\_inclusive\_sentences\_german} under \textit{Apache 2.0 License}, evaluated by accuracy
    \item \textbf{gender\_sensitivity\_chinese}~\cite{China_SOC_2015,China_name_report_2020,China_name_report_2019,China_name_report_2018,China_popular_names_2016,Taiwan_name_analysis_2016} under \textit{Apache 2.0 License}, evaluated by accuracy
    \item \textbf{gender\_sensitivity\_english}~\cite{Bordia2019bias,marcus-etal-1994-penn,Caliskan2017semantics,bolukbasi2016man,rudinger-etal-2018-gender,Lu2020bias,gonen-goldberg-2019-lipstick,hall-maudslay-etal-2019-name,Fellbaum1998wordnet} under \textit{Apache 2.0 License}, evaluated by accuracy
    \item \textbf{general\_knowledge}~\cite{Shane2020questions,Dhingra2017Quasar,rajpurkar-etal-2016-squad,rajpurkar-etal-2018-know,lacker2020gptturing} under \textit{Apache 2.0 License}, evaluated by accuracy
    \item \textbf{geometric\_shapes}~\cite{Bostock2011D3,Marriott2021inclusive,Boillot2019vector} under \textit{Apache 2.0 License}, evaluated by accuracy
    \item \textbf{goal\_step\_wikihow}~\cite{zhang-etal-2020-reasoning} under \textit{Apache 2.0 License}, evaluated by accuracy
    \item \textbf{gre\_reading\_comprehension}~\cite{lai-etal-2017-race} under \textit{Apache 2.0 License}, evaluated by accuracy
    \item \textbf{hhh\_alignment} under \textit{Apache 2.0 License}, evaluated by accuracy
    \item \textbf{high\_low\_game} under \textit{Apache 2.0 License}, evaluated by accuracy
    \item \textbf{hindi\_question\_answering}~\cite{Brown2020,radford2019language,Jain2020indic,lewis-etal-2020-mlqa,artetxe-etal-2020-cross,rajpurkar-etal-2016-squad} under \textit{Apache 2.0 License}, evaluated by accuracy
    \item \textbf{hinglish\_toxicity} under \textit{Apache 2.0 License}, evaluated by accuracy
    \item \textbf{human\_organs\_senses} under \textit{Apache 2.0 License}, evaluated by accuracy
    \item \textbf{hyperbaton}~\cite{Forsyth2014elements} under \textit{Apache 2.0 License}, evaluated by accuracy
    \item \textbf{identify\_math\_theorems}~\cite{Gao2021pile,black2022GPTNeoX,Brown2020,radford2019language,Wankgpt-j} under \textit{Apache 2.0 License}, evaluated by accuracy
    \item \textbf{identify\_odd\_metaphor}~\cite{Lakoff2008metaphors,gao-etal-2018-neural} under \textit{Apache 2.0 License}, evaluated by accuracy
    \item \textbf{implicatures}~\cite{sep-implicature,GEORGE20202316} under \textit{Apache 2.0 License}, evaluated by accuracy
    \item \textbf{implicit\_relations}~\cite{Cain1999inference,Bayat2020relationship,Srivastava2016inferring,lin-etal-2019-reasoning,Massey2015annotating} under \textit{Apache 2.0 License}, evaluated by accuracy
    \item \textbf{intent\_recognition}~\cite{Brown2020,winata2021language,Madotto2020dialogue,Coucke2018snips,madotto2021few} under \textit{Apache 2.0 License}, evaluated by accuracy
    \item \textbf{international\_phonetic\_alphabet\_nli}~\cite{williams-etal-2018-broad} under \textit{Apache 2.0 License}, evaluated by accuracy
    \item \textbf{international\_phonetic\_alphabet\_transliterate}~\cite{Brown2020,Liu2020multilingual,williams-etal-2018-broad} under \textit{Apache 2.0 License}, evaluated by accuracy
    \item \textbf{intersect\_geometry}~\cite{Weston2015toy,Agrawal2015VQA,Trask2018units,Seo_Hajishirzi_Farhadi_Etzioni_2014,hosseini-etal-2014-learning,Polu2020generative,Yang2019theorems} under \textit{Apache 2.0 License}, evaluated by accuracy
    \item \textbf{irony\_identification}~\cite{ZHANG20191633,Ghanem2020irony,SALASZARATE201720} under \textit{Apache 2.0 License}, evaluated by accuracy
    \item \textbf{kanji\_ascii} under \textit{Apache 2.0 License}, evaluated by accuracy
    \item \textbf{kannada}~\cite{Prentice1975joking,Narasimhachar1988kannada,Liu2021synthetic,lev-etal-2004-solving,Lin2021riddlesense} under \textit{Apache 2.0 License}, evaluated by accuracy
    \item \textbf{key\_value\_maps} under \textit{Apache 2.0 License}, evaluated by accuracy
    \item \textbf{language\_games} under \textit{Apache 2.0 License}, evaluated by accuracy
    \item \textbf{linguistic\_mappings}~\cite{mccoy2018revisiting,10.1162/tacl_a_00304,mulligan-etal-2021-structure,RumelhartMcClellandGroup86,10.1162/tacl_a_00247,doi:10.1080/00437956.1958.11659661,BaayenCELEX2} under \textit{Apache 2.0 License}, evaluated by accuracy
    \item \textbf{list\_functions}~\cite{RULE2020900,Rule2020hacker,Green1974progress,Shaw1975inferring,Biermann1978inference,GREEN1981202,Smith1984synthesis,Feser2015synthesizing,Osera2015type,Polikarpova2016program,Cropper2020logic,Graves2014turing,Reed2015neural,Joulin2015patterns,Balog2016deepcoder,Bosnjak2017forth,Gaunt2016terpret,chen2019execution,Kitzelmann2010inductive,Flener2008introduction,Gulwani2017foundations,Devlin2017robust,Ellis2020dream,Cropper2016metagol,Piantadosi2020Fleet} under \textit{Apache 2.0 License}, evaluated by accuracy
    \item \textbf{logical\_args} under \textit{Apache 2.0 License}, evaluated by accuracy
    \item \textbf{logical\_fallacy\_detection}~\cite{Brown2020,hendryckstest2021,10.1145/3442188.3445922,wachsmuth-etal-2017-computational,Yu2020ReClor,Copi2019logic,Oberauer2005effects,Oberauer2000directionality} under \textit{Apache 2.0 License}, evaluated by accuracy
    \item \textbf{logical\_sequence}~\cite{Saxton2019analysing,lin-etal-2020-birds,bowman-dahl-2021-will} under \textit{Apache 2.0 License}, evaluated by accuracy
    \item \textbf{long\_context\_integration} under \textit{Apache 2.0 License}, evaluated by accuracy
    \item \textbf{mathematical\_induction}~\cite{Hendrycks2021mathematical,patel-etal-2021-nlp} under \textit{Apache 2.0 License}, evaluated by accuracy
    \item \textbf{matrixshapes} under \textit{Apache 2.0 License}, evaluated by accuracy
    \item \textbf{metaphor\_boolean}~\cite{Lakoff2008metaphors,bizzoni-lappin-2018-predicting} under \textit{Apache 2.0 License}, evaluated by accuracy
    \item \textbf{metaphor\_understanding}~\cite{Paul1970figurative,tong-etal-2021-recent,radford2019language,Rai2020survey,shutova-2010-automatic,Stowe2020metaphoric,mohler-etal-2016-introducing,shutova-teufel-2010-metaphor,birke-sarkar-2006-clustering,zayed-etal-2020-figure,Steen2010metaphor,Tong2021paraphrasing,bizzoni-lappin-2018-predicting,tsvetkov-etal-2014-metaphor} under \textit{Apache 2.0 License}, evaluated by accuracy
    \item \textbf{minute\_mysteries\_qa}~\cite{Sugawara_Yokono_Aizawa_2017,dunietz-etal-2020-test,kocisky-etal-2018-narrativeqa,https://doi.org/10.48550/arxiv.1604.01696,frermann-etal-2018-whodunnit} under \textit{Apache 2.0 License}, evaluated by accuracy
    \item \textbf{misconceptions}~\cite{Irving2018safety,atanasova-etal-2020-generating-fact,Boller1989fake} under \textit{Apache 2.0 License}, evaluated by accuracy
    \item \textbf{mnist\_ascii} under \textit{Apache 2.0 License}, evaluated by accuracy
    \item \textbf{modified\_arithmetic}~\cite{Brown2020} under \textit{Apache 2.0 License}, evaluated by accuracy
    \item \textbf{moral\_permissibility}~\cite{Hendrycks2020aligning,Lourie2020scruples,Thomson1976killing} under \textit{Apache 2.0 License}, evaluated by accuracy
    \item \textbf{movie\_dialog\_same\_or\_different}~\cite{Park2021diarization} under \textit{Apache 2.0 License}, evaluated by accuracy
    \item \textbf{movie\_recommendation}~\cite{sileo2022reclm,Thorat2015survey,Barkan2016embedding,harper2015movielens} under \textit{Apache 2.0 License}, evaluated by accuracy
    \item \textbf{mult\_data\_wrangling}~\cite{10.1145/3442188.3445922,Tamkin2021understanding,Singh2015predicting,Cropper2016meta,Wu2012rules,Gulwani2015realworld,Contreras2018general,Contreras2020automated,PetrovaAntonova2020cleaning,Huynh2012openrefine,Kandel2011wrangler,Bhupatiraju2017deep,ijcai2017-227,Gulwani2012spreadsheet,Gulwani2011automating,Singh2016transforming} under \textit{Apache 2.0 License}, evaluated by accuracy
    \item \textbf{multiemo}~\cite{10.1007/978-3-030-77964-1_24} under \textit{Apache 2.0 License}, evaluated by accuracy
    \item \textbf{multistep\_arithmetic}~\cite{Flanagan2014cattell} under \textit{Apache 2.0 License}, evaluated by accuracy
    \item \textbf{muslim\_violence\_bias}~\cite{Abid2021persistent,10.1145/3442188.3445922} under \textit{Apache 2.0 License}, evaluated by accuracy
    \item \textbf{natural\_instructions}~\cite{mishra2021crosstask} under \textit{Apache 2.0 License}, evaluated by accuracy
    \item \textbf{navigate}~\cite{Graves2016hybrid,Henaff2016tracking,Geva2020transformer,Chen2019touchdown,kryscinski-etal-2020-evaluating,Cote2018TextWorld,Luketina2019ASO,thawani-etal-2021-representing,Lake2017generalization} under \textit{Apache 2.0 License}, evaluated by accuracy
    \item \textbf{nonsense\_words\_grammar} under \textit{Apache 2.0 License}, evaluated by accuracy
    \item \textbf{object\_counting}~\cite{doi:10.1126/science.aaa1379,Wang2019objects,Zhang2018objects,Brown2020} under \textit{Apache 2.0 License}, evaluated by accuracy
    \item \textbf{odd\_one\_out}~\cite{Resnik1995taxonomy,Resnik1999semantic,jiang-conrath-1997-semantic,Li2003similarity,Banerjee2003gloss,Jarmasz2012Roget,hughes-ramage-2007-lexical,Tsatsaronis2010thesaurus,Tsatsaronis2009omiotis,morris-hirst-1991-lexical,Strube2006wiki,Ponzetto2007wiki,Gabrilovich2007wiki,Milne2008effective,yeh-etal-2009-wikiwalk,Radinsky2011word,Cilibrasi2007google,Deerwester1990indexing,reisinger-mooney-2010-multi,ElYaniv2013semantic} under \textit{Apache 2.0 License}, evaluated by accuracy
    \item \textbf{paragraph\_segmentation} under \textit{Apache 2.0 License}, evaluated by accuracy
    \item \textbf{parsinlu\_qa}~\cite{Khashabi2020parsinlu} under \textit{Apache 2.0 License}, evaluated by accuracy
    \item \textbf{penguins\_in\_a\_table}~\cite{herzig-etal-2020-tapas} under \textit{Apache 2.0 License}, evaluated by accuracy
    \item \textbf{periodic\_elements} under \textit{Apache 2.0 License}, evaluated by accuracy
    \item \textbf{persian\_idioms} under \textit{Apache 2.0 License}, evaluated by accuracy
    \item \textbf{phrase\_relatedness}~\cite{asaadi-etal-2019-big,levy-etal-2015-tr9856,EinDor2018semantic} under \textit{Apache 2.0 License}, evaluated by accuracy
    \item \textbf{physical\_intuition} under \textit{Apache 2.0 License}, evaluated by accuracy
    \item \textbf{physics} under \textit{Apache 2.0 License}, evaluated by accuracy
    \item \textbf{physics\_questions}~\cite{Ling2017induction,amini-etal-2019-mathqa} under \textit{Apache 2.0 License}, evaluated by accuracy
    \item \textbf{polish\_sequence\_labeling}~\cite{Nguyen2007comparison,Rei2017sequence,Gu2018universal} under \textit{Apache 2.0 License}, evaluated by accuracy
    \item \textbf{presuppositions\_as\_nli}~\cite{Heim1983projection,deMarneffe_Simons_Tonhauser_2019,white-etal-2018-lexicosyntactic,jeretic-etal-2020-natural} under \textit{Apache 2.0 License}, evaluated by accuracy
    \item \textbf{program\_synthesis}~\cite{Gulwani2017synthesis} under \textit{Apache 2.0 License}, evaluated by accuracy
    \item \textbf{protein\_interacting\_sites}~\cite{Dhole2014sequence,Singh2014springs,10.1093/bioinformatics/btaa750,10.1093/bioinformatics/btq302} under \textit{Apache 2.0 License}, evaluated by accuracy
    \item \textbf{python\_programming\_challenge}~\cite{Allamanis2018survey,pmlr-v119-alon20a,hendrycks2021measuring} under \textit{Apache 2.0 License}, evaluated by accuracy
    \item \textbf{qa\_wikidata}~\cite{radford2019language,47761,Weber2011websearch} under \textit{Apache 2.0 License}, evaluated by accuracy
    \item \textbf{question\_answer\_creation} under \textit{Apache 2.0 License}, evaluated by accuracy
    \item \textbf{question\_selection}~\cite{rajpurkar-etal-2016-squad} under \textit{Apache 2.0 License}, evaluated by accuracy
    \item \textbf{real\_or\_fake\_text}~\cite{dugan-etal-2020-roft,ippolito-etal-2020-automatic,Solaiman2019release,Zellers2019defending,Brown2020,Bakhtin2019fake,Sandhaus2008times,fan-etal-2018-hierarchical,Marin2021recipe} under \textit{Apache 2.0 License}, evaluated by accuracy
    \item \textbf{reasoning\_about\_colored\_objects}~\cite{Hendricks2018snowboard,hosseini-etal-2014-learning,WINOGRAD19721,wang-etal-2016-learning-language,jayannavar-etal-2020-learning,suhr-etal-2019-executing,Thomason2015commands,mitchell-etal-2010-natural,viethen-dale-2008-use,gatt-etal-2009-tuna,mitchell-etal-2013-generating,liang-etal-2018-meaning} under \textit{Apache 2.0 License}, evaluated by accuracy
    \item \textbf{rephrase} under \textit{Apache 2.0 License}, evaluated by accuracy
    \item \textbf{riddle\_sense}~\cite{Lin2021riddlesense,talmor-etal-2019-commonsenseqa} under \textit{Apache 2.0 License}, evaluated by accuracy
    \item \textbf{roots\_optimization\_and\_games}~\cite{Lample2019deep,Polu2020generative,Amos2017optnet,Agrawal2020convex} under \textit{Apache 2.0 License}, evaluated by accuracy
    \item \textbf{ruin\_names}~\cite{HumorinLanguage,He2020convex,amin-burghardt-2020-survey,Annamoradnejad2020colbert,blinov-etal-2019-large,Yan2017funny,Frolovs2019humor} under \textit{Apache 2.0 License}, evaluated by accuracy
    \item \textbf{salient\_translation\_error\_detection} under \textit{Apache 2.0 License}, evaluated by accuracy
    \item \textbf{scientific\_press\_release} under \textit{Apache 2.0 License}, evaluated by accuracy
    \item \textbf{self\_awareness}~\cite{Yudkowsky2008global,Chella2020robots,Schick2021self,Kounev2017self,Huttunen2017self,Wallace2020phone,Clark1994spider,HOROWITZ201717,Branwen2021fiction,Chu2017steganography} under \textit{Apache 2.0 License}, evaluated by accuracy
    \item \textbf{self\_evaluation\_courtroom}~\cite{Hildebrandt2018regulation,Daley2021Chicago,king-cook-2020-evaluating} under \textit{Apache 2.0 License}, evaluated by accuracy
    \item \textbf{self\_evaluation\_tutoring}~\cite{Zhang2019agent,Irving2018safety} under \textit{Apache 2.0 License}, evaluated by accuracy
    \item \textbf{semantic\_parsing\_in\_context\_sparc}~\cite{yu-etal-2019-sparc,yu-etal-2018-spider,yu-etal-2019-cosql} under \textit{Apache 2.0 License}, evaluated by accuracy
    \item \textbf{semantic\_parsing\_spider}~\cite{yu-etal-2018-spider,yu-etal-2019-sparc,yu-etal-2019-cosql} under \textit{Apache 2.0 License}, evaluated by accuracy
    \item \textbf{sentence\_ambiguity} under \textit{Apache 2.0 License}, evaluated by accuracy
    \item \textbf{similarities\_abstraction}~\cite{Nasreddine2005moca,Wechsler2008scale} under \textit{Apache 2.0 License}, evaluated by accuracy
    \item \textbf{simp\_turing\_concept}~\cite{Bohm1964turing,Sun2020evolution,Devlin2018BERT,radford2019language,Brown2020,Vaswani2017attention,hendryckstest2021,xu-etal-2020-autoqa,Izacard2020leveraging,Zhu_2015teaching,CAKMAK2014198,Goodman2016pragmatic,Degen2020redundancy,NIPS2011_f9028fae,Basu_Christensen_2013,Yang2017explainable,ijcai2018-356,Telle2019teaching,Chater2003simplicity,Hupkes2020decomposed,lakretz-etal-2019-emergence,Toshniwal2021blindfolded,bender2020climbing,Kuhl2020human,Marcus2020bloviator,Sinha2019CLUTRR,McClelland2019extending} under \textit{Apache 2.0 License}, evaluated by accuracy
    \item \textbf{simple\_ethical\_questions}~\cite{Hendrycks2020aligning,Lourie2020scruples} under \textit{Apache 2.0 License}, evaluated by accuracy
    \item \textbf{simple\_text\_editing}~\cite{Branwen2021fiction,Malmi2019encode,Faltings2020command} under \textit{Apache 2.0 License}, evaluated by accuracy
    \item \textbf{snarks}~\cite{Brown2020,Devlin2018BERT,Lan2019albert,Liu2019roberta,radford2019language,Annamoradnejad2020colbert,chen-soo-2018-humor,Mao2019bert,weller-seppi-2019-humor,Khodak2017sarcasm,Ghosh2020sarcasm,gonzalez-ibanez-etal-2011-identifying,joshi-etal-2015-harnessing,McCoy2019right,Kaushik2019learning,Gardner2020boundaries,Sennrich2016grammatical,burlot-etal-2018-wmt18,Naik2018stress,zhang-etal-2016-tweet,Felbo2017emoji,Pant2020sarcasm,Pelser2019deep} under \textit{Apache 2.0 License}, evaluated by accuracy
    \item \textbf{social\_support}~\cite{wang-jurgens-2018-going} under \textit{Apache 2.0 License}, evaluated by accuracy
    \item \textbf{social\_iqa}~\cite{sap-etal-2019-social,Bisk_Zellers_Lebras_Gao_Choi_2020,talmor-etal-2019-commonsenseqa,Zellers2018recognition} under \textit{Apache 2.0 License}, evaluated by accuracy
    \item \textbf{spelling\_bee}~\cite{DBLP:journals/corr/Ginsberg14} under \textit{Apache 2.0 License}, evaluated by accuracy
    \item \textbf{sports\_understanding} under \textit{Apache 2.0 License}, evaluated by accuracy
    \item \textbf{squad\_shifts}~\cite{pmlr-v119-miller20a,Brown2020,rajpurkar-etal-2016-squad,Baumgartner2020pushshift,McAuley2015image} under \textit{Apache 2.0 License}, evaluated by accuracy
    \item \textbf{subject\_verb\_agreement}~\cite{LAKRETZ2021104699,lakretz-etal-2019-emergence,10.1162/tacl_a_00115,gulordava-etal-2018-colorless,marvin-linzen-2018-targeted,Goldberg2019bert,https://doi.org/10.48550/arxiv.2101.02258,Wolf2019bert} under \textit{Apache 2.0 License}, evaluated by accuracy
    \item \textbf{sudoku}~\cite{Wang2019satnet,Russell2002modern,Garcez2020neurosymbolic,Huang2018gamepad,Hendrycks2021mathematical} under \textit{Apache 2.0 License}, evaluated by accuracy
    \item \textbf{sufficient\_information} under \textit{Apache 2.0 License}, evaluated by accuracy
    \item \textbf{suicide\_risk}~\cite{Gaur2019suicide,mohammadi-etal-2019-clac-clpsych,matero-etal-2019-suicide,shing-etal-2018-expert} under \textit{Apache 2.0 License}, evaluated by accuracy
    \item \textbf{swahili\_english\_proverbs} under \textit{Apache 2.0 License}, evaluated by accuracy
    \item \textbf{swedish\_to\_german\_proverbs}~\cite{Hanzen2007Rome,Korhonen2009,Meister2007,Mieder2019andere} under \textit{Apache 2.0 License}, evaluated by accuracy
    \item \textbf{taboo}~\cite{2017arXivtriviaqa} under \textit{Apache 2.0 License}, evaluated by accuracy
    \item \textbf{talkdown}~\cite{wang-potts-2019-talkdown,Mendelsohn2020framework,Fiske1993controlling,Nolan2013things,breitfeller-etal-2019-finding,perez-almendros-etal-2020-dont} under \textit{Apache 2.0 License}, evaluated by accuracy
    \item \textbf{temporal\_sequences}~\cite{Elazar2021measuring,Pustejovsky2004timeml,Sanampudi2010temporal,Han2020econet,Ma2021eventplus,Brown2020,Petroni2019bases} under \textit{Apache 2.0 License}, evaluated by accuracy
    \item \textbf{tense}~\cite{Logeswaraw2018content} under \textit{Apache 2.0 License}, evaluated by accuracy
    \item \textbf{text\_navigation\_game}~\cite{Vinyals2019starcraft,Kuttler2020nethack,Kanagawa2019rogue,noever2020chess} under \textit{Apache 2.0 License}, evaluated by accuracy
    \item \textbf{timedial}~\cite{qin-etal-2021-timedial,li-etal-2017-dailydialog,Zhou2019vacation} under \textit{Apache 2.0 License}, evaluated by accuracy
    \item \textbf{topical\_chat}~\cite{gopalakrishnan19_interspeech,mehri-eskenazi-2020-usr,gopalakrishnan20_interspeech,hedayatnia-etal-2020-policy} under \textit{Apache 2.0 License}, evaluated by accuracy
    \item \textbf{tracking\_shuffled\_objects}~\cite{Liu2020interpretable,Dong2020syllogism} under \textit{Apache 2.0 License}, evaluated by accuracy
    \item \textbf{training\_on\_test\_set} under \textit{Apache 2.0 License}, evaluated by accuracy
    \item \textbf{truthful\_qa}~\cite{Brown2020,Sellam2020bleurt,Amodei2016concrete,Leike2018scalable,kenton2021alignment,Clark2018ARC,Bhakthavatsalam2021solved,hendryckstest2021,khashabi-etal-2020-unifiedqa,Kreps2020news,Maynez2020faithfulness,Gabriel2020figure,wang-etal-2020-asking,Stiennon2020summarize,Lewis2020retrieval,Krishna2021hurdles,gehrmann-etal-2021-gem,Xu2020Recipes,Dinan2019queens,Tamkin2021understanding,bowman-dahl-2021-will} under \textit{Apache 2.0 License}, evaluated by accuracy
    \item \textbf{twenty\_questions}~\cite{rajpurkar-etal-2016-squad,Choi2018quac,reddy-etal-2019-coqa,Aliannejadi2019asking,Clark2019boolq,Zhang2019agent} under \textit{Apache 2.0 License}, evaluated by accuracy
    \item \textbf{understanding\_fables}~\cite{Reimers2019sentence,salazar-etal-2020-masked,Wolf2019huggingface} under \textit{Apache 2.0 License}, evaluated by accuracy
    \item \textbf{undo\_permutation}~\cite{Pham2020order} under \textit{Apache 2.0 License}, evaluated by accuracy
    \item \textbf{unit\_conversion}~\cite{Hendrycks2021mathematical,geva-etal-2020-injecting} under \textit{Apache 2.0 License}, evaluated by accuracy
    \item \textbf{unit\_interpretation} under \textit{Apache 2.0 License}, evaluated by accuracy
    \item \textbf{unnatural\_in\_context\_learning}~\cite{Brown2020,Kaplan2020scaling,Henighan2020scaling,Hernandez2021scaling,bahri2021explaining,Wang2019SuperGLUEAS,Hernandez_2020,Hendrycks2021mathematical,hendrycks2021measuring,Liu2021good,Zhao2021calibrate,Perez2021true} under \textit{Apache 2.0 License}, evaluated by accuracy
    \item \textbf{unqover}~\cite{li-etal-2020-unqovering,Caliskan2017semantics,rudinger-etal-2018-gender,zhao-etal-2018-gender,Dev_Li_Phillips_Srikumar_2020,stanovsky-etal-2019-evaluating,Nadeem2020StereoSet,sheng-etal-2019-woman,Zhang2020hurtful} under \textit{Apache 2.0 License}, evaluated by accuracy
    \item \textbf{web\_of\_lies} under \textit{Apache 2.0 License}, evaluated by accuracy
    \item \textbf{what\_is\_the\_tao} under \textit{Apache 2.0 License}, evaluated by accuracy
    \item \textbf{which\_wiki\_edit} under \textit{Apache 2.0 License}, evaluated by accuracy
    \item \textbf{word\_problems\_on\_sets\_and\_graphs}~\cite{Bency2019path,Mnih2013atari,Russell2002modern,Besold2017survey,Clark2020transformers,wang-etal-2018-glue,lacker2020gptturing} under \textit{Apache 2.0 License}, evaluated by accuracy
    \item \textbf{word\_sorting}~\cite{Grover2019stochastic} under \textit{Apache 2.0 License}, evaluated by accuracy
    \item \textbf{word\_unscrambling}~\cite{nishino-etal-2019-generating,Rozner2021cryptic,Jones2020robust,MAYS1991517,Edizel2019misspelling,Sakaguchi2016character,Kim2015character,Xue2021token,Wu2020applying,Rust2020tokenizer} under \textit{Apache 2.0 License}, evaluated by accuracy
    \item \textbf{yes\_no\_black\_white} under \textit{Apache 2.0 License}, evaluated by accuracy
\end{itemize}

\subsection{\newtxt{BigBench-Lite Datasets}}

\begin{itemize}
    \item \textbf{auto\_debugging}~\cite{Zaremba2014learning} under \textit{Apache 2.0 License}, evaluated by accuracy
    \item \textbf{bbq\_lite\_json}~\cite{Crawford2017trouble,khashabi-etal-2020-unifiedqa,li-etal-2020-unqovering} under \textit{Apache 2.0 License}, evaluated by accuracy
    \item \textbf{code\_line\_description}~\cite{Alon2018code2seq} under \textit{Apache 2.0 License}, evaluated by accuracy
    \item \textbf{conceptual\_combinations}~\cite{Fodor1975,FODOR19883,smolensky_1988,lake_ullman_tenenbaum_gershman_2017,Lake2020wordmeaning,Marcus2020next,henrich_heine_norenzayan_2010,Murphy1988complex} under \textit{Apache 2.0 License}, evaluated by accuracy
    \item \textbf{conlang\_translation}~\cite{Canfield2010Klingon,sahin-etal-2020-puzzling,sennrich-zhang-2019-revisiting} under \textit{Apache 2.0 License}, evaluated by accuracy
    \item \textbf{emoji\_movie}~\cite{Cruse2020emoji,Instagram2015emojineering,ChandraGuntuku_Li_Tay_Ungar_2019,eisner-etal-2016-emoji2vec,Mayne2020emoji,Boillot2019vector} under \textit{Apache 2.0 License}, evaluated by accuracy
    \item \textbf{formal\_fallacies\_syllogisms\_negation}~\cite{Kassner2019probes,Talmor2019olmpics,Betz2020critical} under \textit{Apache 2.0 License}, evaluated by accuracy
    \item \textbf{hindu\_knowledge} under \textit{Apache 2.0 License}, evaluated by accuracy
    \item \textbf{known\_unknowns}~\cite{Liu2021token,Xiao2021hallucination,Shuster2021retrieval,Zhou2020detecting,Dziri2021pathhunter} under \textit{Apache 2.0 License}, evaluated by accuracy
    \item \textbf{language\_identification}~\cite{brown-2014-non} under \textit{Apache 2.0 License}, evaluated by accuracy
    \item \textbf{linguistics\_puzzles}~\cite{bozhanov-derzhanski-2013-rosetta,radev-etal-2008-north,sennrich-zhang-2019-revisiting,Clark2018ARC,sahin-etal-2020-puzzling} under \textit{Apache 2.0 License}, evaluated by accuracy
    \item \textbf{logic\_grid\_puzzle} under \textit{Apache 2.0 License}, evaluated by accuracy
    \item \textbf{logical\_deduction} under \textit{Apache 2.0 License}, evaluated by accuracy
    \item \textbf{misconceptions\_russian}~\cite{thorne-etal-2018-fever,lee-etal-2020-language} under \textit{Apache 2.0 License}, evaluated by accuracy
    \item \textbf{novel\_concepts}~\cite{Santoro2021symbolic} under \textit{Apache 2.0 License}, evaluated by accuracy
    \item \textbf{operators}~\cite{Brown2020,Kassner2020pretrained,Hendrycks2021mathematical,Saxton2019analysing} under \textit{Apache 2.0 License}, evaluated by accuracy
    \item \textbf{parsinlu\_reading\_comprehension}~\cite{Khashabi2020parsinlu,xue-etal-2021-mt5,rajpurkar-etal-2016-squad} under \textit{Apache 2.0 License}, evaluated by accuracy
    \item \textbf{play\_dialog\_same\_or\_different}~\cite{Park2021diarization} under \textit{Apache 2.0 License}, evaluated by accuracy
    \item \textbf{repeat\_copy\_logic}~\cite{Graves2014turing} under \textit{Apache 2.0 License}, evaluated by accuracy
    \item \textbf{strange\_stories}~\cite{Happe1994test,White2009autism} under \textit{Apache 2.0 License}, evaluated by accuracy
    \item \textbf{strategyqa}~\cite{Geva2021Aristotle} under \textit{Apache 2.0 License}, evaluated by accuracy
    \item \textbf{symbol\_interpretation}~\cite{Brown2020,Johnson2016CLEVR,Santoro2017simple,Hudson2019gqa,Sennrich2015rare} under \textit{Apache 2.0 License}, evaluated by accuracy
    \item \textbf{vitaminc\_fact\_verification}~\cite{schuster-etal-2021-get} under \textit{Apache 2.0 License}, evaluated by accuracy
    \item \textbf{winowhy}~\cite{zhang-etal-2020-winowhy,Devlin2018BERT,Liu2019roberta,kocijan-etal-2019-surprisingly,rahman-ng-2012-resolving,Sakaguchi2020WINOGRANDE} under \textit{Apache 2.0 License}, evaluated by accuracy
\end{itemize}

\section{Efficiency Analysis}
\method{} introduces minimal computational overhead by requiring only two lightweight operations per LoRA layer: a single cosine similarity calculation between the query's task embedding and global routing vectors and a simple vector addition to combine this with the local routing score. With typical values for $N$ (experts) and $d_g$ (embedding dimension) in the hundreds, this amounts to just $(N\times d_g + d_g)$ FLOPs per layer, which is negligible compared to the base model's computation.

\end{document}